\documentclass[lettersize,journal]{IEEEtran}
\usepackage{amsmath,amsfonts}
\usepackage{algorithmic}
\usepackage{algorithm}
\usepackage{array}
\usepackage[caption=false,font=normalsize,labelfont=sf,textfont=sf]{subfig}
\usepackage{textcomp}
\usepackage{stfloats}
\usepackage{url}
\usepackage{verbatim}
\usepackage{graphicx}
\usepackage{cite}
\usepackage{multirow}
\usepackage{booktabs}
\usepackage{graphicx}
\usepackage{adjustbox}
\usepackage{overpic}
\usepackage{colortbl}
\usepackage{xcolor}
\usepackage{makecell}
\usepackage{tabularx}

\begin{document}

\title{SCAWaveNet: A Spatial-Channel Attention-Based Network for Global Significant Wave Height Retrieval}

\author{Chong Zhang, Xichao Liu, Yibing Zhan,~\IEEEmembership{Member,~IEEE}, Dapeng Tao,~\IEEEmembership{Member,~IEEE}, Jun Ni, Jinwei Bu,~\IEEEmembership{Member,~IEEE}
\thanks{This work was supported in part by the National Natural Science Foundation of China under Grants $62302429$ and $42404037$, and in part by the Xingdian Talents Support Project of Yunnan Province.(\emph{Corresponding author: Jun Ni \& Jinwei Bu})}

\thanks{C. Zhang, X. Liu, D. Tao, and J. Ni are with the School of Information Science \& Engineering, Yunnan University, Kunming, China.}
\thanks{Y. Zhan is with Yunnan United Vision Technology Co., Ltd, Kunming, China.}
\thanks{J. Bu is with the Faculty of Land Resources Engineering, Kunming University of Science and Technology, Kunming, China.}
}


\maketitle

\begin{abstract}
Recent advancements in spaceborne GNSS missions have produced extensive global datasets, providing a robust basis for deep learning-based significant wave height (SWH) retrieval. While existing deep learning models predominantly utilize CYGNSS data with four-channel information, they often adopt single-channel inputs or simple channel concatenation without leveraging the benefits of cross-channel information interaction during training. To address this limitation, a novel spatial–channel attention-based network, namely SCAWaveNet, is proposed for SWH retrieval. Specifically, features from each channel of the DDMs are modeled as independent attention heads, enabling the fusion of spatial and channel-wise information. For auxiliary parameters, a lightweight attention mechanism is designed to assign weights along the spatial and channel dimensions. The final feature integrates both spatial and channel-level characteristics. Model performance is evaluated using four-channel CYGNSS data. When ERA5 is used as a reference, SCAWaveNet achieves an average RMSE of 0.438 m. When using buoy data from NDBC, the average RMSE reaches 0.432 m. Compared to state-of-the-art models, SCAWaveNet reduces the average RMSE by at least 3.52\% on the ERA5 dataset and by 5.68\% on the NDBC buoy observations. The code is available at https://github.com/Clifx9908/SCAWaveNet.

\end{abstract}


\section{Introduction} \label{Introduction}
\IEEEPARstart{A}{s} a critical indicator of ocean wave energy and sea state conditions, Significant Wave Height (SWH) plays a key role in marine engineering, maritime safety, and climate change research \cite{yu2025comparative}. Accurate retrieval of SWH provides a fundamental reference for the design of offshore structures, effectively mitigating the risk of structural failure under extreme sea conditions \cite{haselsteiner2020predicting}. Precise SWH data also optimizes maritime route planning, enhancing navigation safety and fuel efficiency \cite{mannarini2019graph}. In addition, long-term SWH monitoring aids in identifying extreme events and assessing marine resources, supporting sustainable ocean development \cite{ewans2023uncertainties, patra2021changes}. 

In recent years, Spaceborne Global Navigation Satellite System Reflectometry (GNSS-R) has become a mainstream technology for SWH retrieval \cite{qin2021significant}. Key missions include TechDemoSat-1 \cite{unwin2016spaceborne}, Cyclone GNSS (CYGNSS) \cite{ruf2016new}, SMAP-R \cite{carreno2017spaceborne}, Fsscat \cite{munoz2020orbit}, Spire \cite{setti2023evaluation}, Fengyun-3E/3F/3G \cite{sun2023gnos}, BuFeng-1 A/B \cite{jing2024review}, and Tianmu-1 \cite{bu2024land, liu2025performance}. Unlike traditional methods \cite{bai2025retrieval, wang2025range, quach2020deep}, GNSS-R enables global, high-resolution SWH observations without relying on costly in situ instruments, substantially reducing operational costs. It remains robust under extreme sea conditions, providing reliable data in various weather and illumination scenarios \cite{chen2025diffwater, hu2025gnss, du2024deep, rodriguez2023latest}. This technique improves the accuracy of SWH retrieval and expands the scope of scientific and operational marine applications.

\begin{figure*}[!ht]
    \centering
    \includegraphics[width=\linewidth]{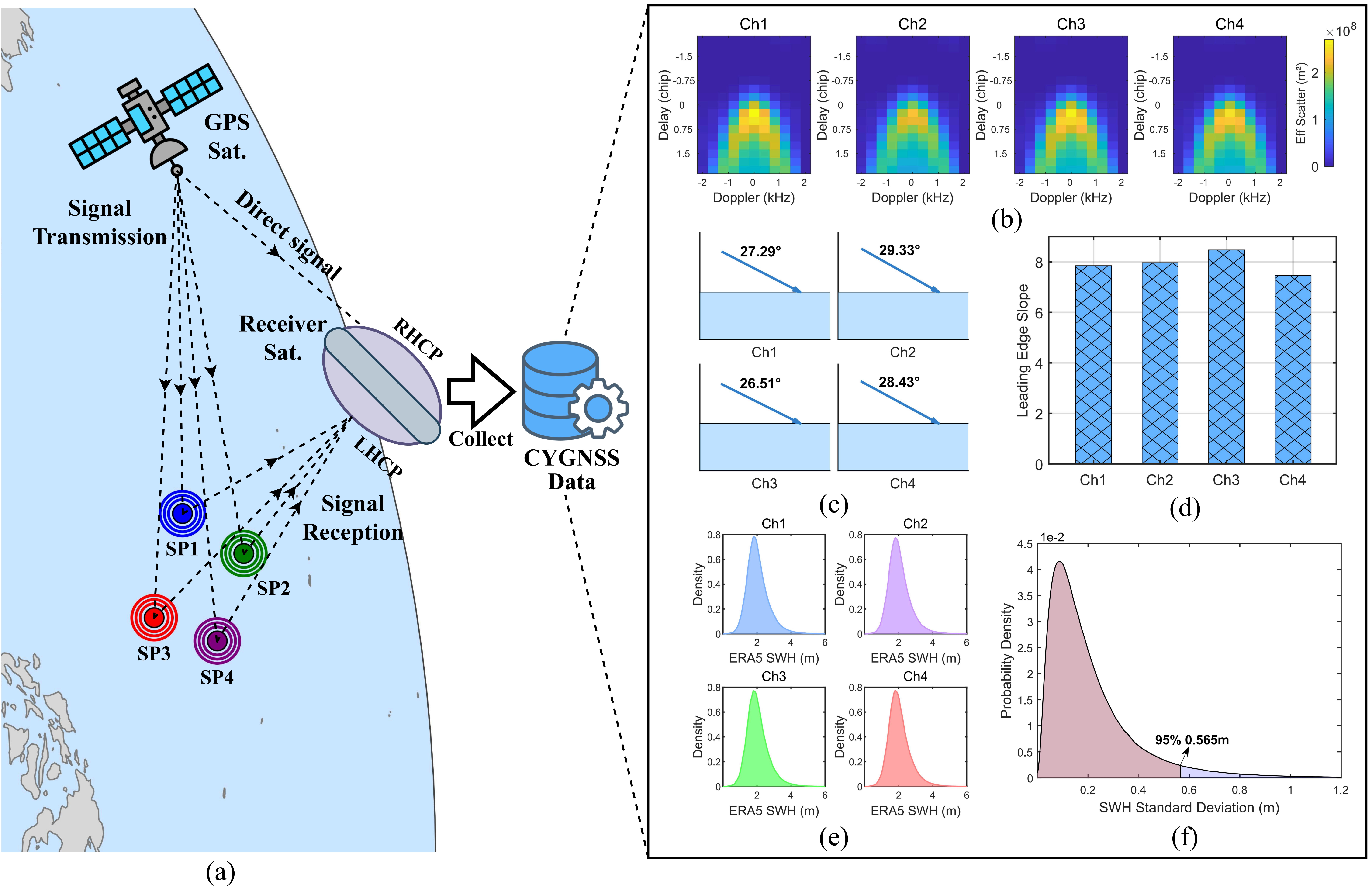}
    \caption{Methodology of CYGNSS data collection.}
    \label{fig:GNSS-4chs}
\end{figure*}

As a powerful tool in GNSS-R observation, machine learning (ML) techniques have been widely applied to SWH retrieval \cite{bu2025machine, sithara2025machine, jin2024remote}. Traditional ML approaches, such as Bagging Tree (BT) \cite{wang2022significant}, Random Forest (RF) \cite{bu2024ocean} and eXtreme Gradient Boosting (XGBoost) \cite{han2024significant}, typically construct regression models based on auxiliary parameters (APs). While these models demonstrate promising performance, they neglect the valuable information in two-dimensional Delay-Doppler maps (DDMs), thus constraining SWH retrieval accuracy.

Recent advancements in deep learning (DL) models have enabled high-resolution global SWH retrieval through architectures like Multi-scale Conv-BiLSTM \cite{wang2024ocean}, CNN-ConvLSTM-FCN \cite{bu2024estimating}, WaveTransNet \cite{qiao2024wavetransnet}, and ViT-Wave \cite{zhou2024enhancing}. In these methods, dual-branch architectures are widely adopted, where Convolutional Neural Networks (CNNs) extract features from DDMs in parallel with Multilayer Perceptrons (MLPs) processing APs. However, the fixed-weight nature of CNNs and MLPs hinders their ability to adjust features in various sea conditions, and they are limited in capturing long-range spatial dependencies. Inspired by Transformer \cite{vaswani2017attention} and Vision Transformer (ViT) \cite{dosovitskiy2021an}, recent studies have incorporated attention mechanisms to enhance spatial feature flexibility and dependency modeling. Yet, most methods either rely on single-channel CYGNSS data or naively aggregate all channels, overlooking inherent channel-wise information and thus restricting architectural optimization potential.

CYGNSS data inherently contain four-channel information. As depicted in Fig.~\ref{fig:GNSS-4chs}(a), each receiver satellite in the CYGNSS constellation carries a four-channel Delay-Doppler Mapping Instrument (DDMI), capable of simultaneously capturing direct GPS signals and four reflected signals \cite{ruf2022cygnss, wang2025measuring, ruf2018new}. The direct signal enables precise receiver localization, the reflected signals capture sea surface roughness for SWH estimation. During signal propagation, the specular point locations of each channel are spatially close, leveraging this spatial similarity can improve the model’s predictive performance \cite{zhang2021integrating}. In addition, CYGNSS datasets include diverse DDMs (e.g., effective scattering area) and APs (e.g., specular point incidence angle, leading edge slope). Notably, the four channels exhibit similarity in DDM/AP characteristics due to their aligned spatial coverage and synchronous acquisition (Fig.~\ref{fig:GNSS-4chs}(b)-(d)). 

To further quantify the intra-channel similarity, we analyzed the SWH density distribution derived from CYGNSS-ERA5 data in 2022, along with the standard deviation (SD) distribution across channels. As shown in Fig.~\ref{fig:GNSS-4chs}(e), the SWH distributions are highly consistent among channels, with most data concentrated between 1-3 m. Furthermore, Fig.~\ref{fig:GNSS-4chs}(f) reveals that the 95th percentile of the SWH SD is 0.565 m, indicating minimal inter-channel variability for the majority samples. These findings confirm the similarity of reference values across CYGNSS channels, motivating the development of channel-aware modeling approaches.

Inspired by CYGNSS's multi-channel observation mechanism, we propose SCAWaveNet, a \textbf{S}patial–\textbf{C}hannel \textbf{A}ttention (SCA) based model for SWH retrieval. Unlike prior works, SCAWaveNet introduces a novel four-channel modeling paradigm by jointly processing DDMs and APs from all channels. First, an SCA-based Transformer Encoder is employed to extract DDM features. Specifically, the DDMs are embedded, and the embeddings from each channel are treated as independent attention heads in multi-head self-attention block, enabling cross-dimensional information fusion. In parallel, a lightweight dual-dimensional attention mechanism is applied to generate spatial-channel weights for APs, and produce channel-aware features. Finally, these two types of features are fused and passed through a task-specific head to generate SWH estimates for all channels. 

In summary, the contributions of this work are as follows:

\begin{itemize}
    \item We propose SCAWaveNet, which dynamically adapts features across spatial and channel dimensions. To our best knowledge, this is the first application of spatial-channel attention mechanisms in GNSS-R SWH retrieval.

    \item We design a data processing framework for four-channel CYGNSS data and conduct a comprehensive evaluation of four-channel information utilization for SWH retrieval. Experiments on the ERA5 and NDBC buoy datasets validate the effectiveness and robustness of SCAWaveNet.
        
    \item We analyze architectural differences between single-channel and four-channel models, investigating the impact of channel independence/dependence strategies on retrieval accuracy. Additional ablation studies further clarify the role of key design choices in SCAWaveNet.

\end{itemize}

\section{Related Work} \label{Related Work}
In this section, we present a brief review of machine learning-based SWH retrieval models and introduce channel strategies in deep learning, emphasizing the significance of employing a spatial-channel attention mechanism in SWH retrieval.

\subsection{Machine Learning Approaches for SWH Retrieval Task}

With the rapid development of ML technologies, various methods have been applied in GNSS-R observations to obtain ocean parameters. Many researchers have developed models capable of retrieving global SWH with strong performance and generalizability \cite{bu2025machine}. Early ML models primarily relied on APs for modeling. Wang et al. \cite{wang2022significant} developed five multivariate regression models using 12 APs, with the BT model achieving the best performance (0.48 m RMSE on ERA5 and 0.44 m on NDBC). Bu et al. \cite{bu2024ocean} constructed five tree-based models, where the RF model exhibited superior performance (0.411 m RMSE in 0–8 m). To improve performance under high SWH conditions, Han et al. \cite{han2024significant} proposed XGBoost-SC, a refined XGBoost model, which achieved 0.94 m RMSE for 3–6 m SWH and 2.79 m RMSE for SWH \textgreater4 m. Although these models perform well, they can only extract features from one-dimensional data and cannot leverage information from two-dimensional DDMs, constraining the performance. 

In recent years, DL architectures such as MLP, CNN, and Transformer have gained popularity due to their superior spatial modeling capabilities and efficiency in handling high-dimensional data. Bu et al. developed models including Multi-scale Conv-BiLSTM \cite{wang2024ocean}, CNN-ConvLSTM-FCN \cite{bu2024estimating}, DCNN \cite{bu2023combining}, and GloWH-Net \cite{jinwei53deep}, which demonstrated enhanced generalizability in global SWH retrieval. These models share a common dual-branch architecture using CNNs for DDM feature extraction and MLPs for AP processing, but differ in the network structures of individual branches. However, these models have limited receptive fields to capture long-range dependencies, hindering the capture of complex features, while static weight parameters prevent dynamic adaptation to varying sea conditions.

Based on the Transformer and ViT architectures, researchers proposed WaveTransNet \cite{qiao2024wavetransnet} and ViT-Wave \cite{zhou2024enhancing}, incorporating attention mechanisms. Both apply Transformer encoders to DDM patch embeddings but differ in AP branch design: ViT-Wave directly concatenates raw APs with DDM features, while WaveTransNet uses attention-based AP feature extraction before two-branch fusion. These attention mechanisms effectively capture long-range dependencies and enhance dynamic feature modeling. However, these models remain ``single-channel'' in paradigm, relying on either single-channel data or merged multi-channel inputs to predict a single SWH value. This approach neglects channel-wise information inherent in multi-channel GNSS-R data. To resolve this issue, channel-aware modeling could optimize architectures, improve retrieval accuracy, and expand applications.

\subsection{Channel Strategies in Deep Learning}

\begin{figure}[!ht]
    \centering
    \includegraphics[width=\linewidth]{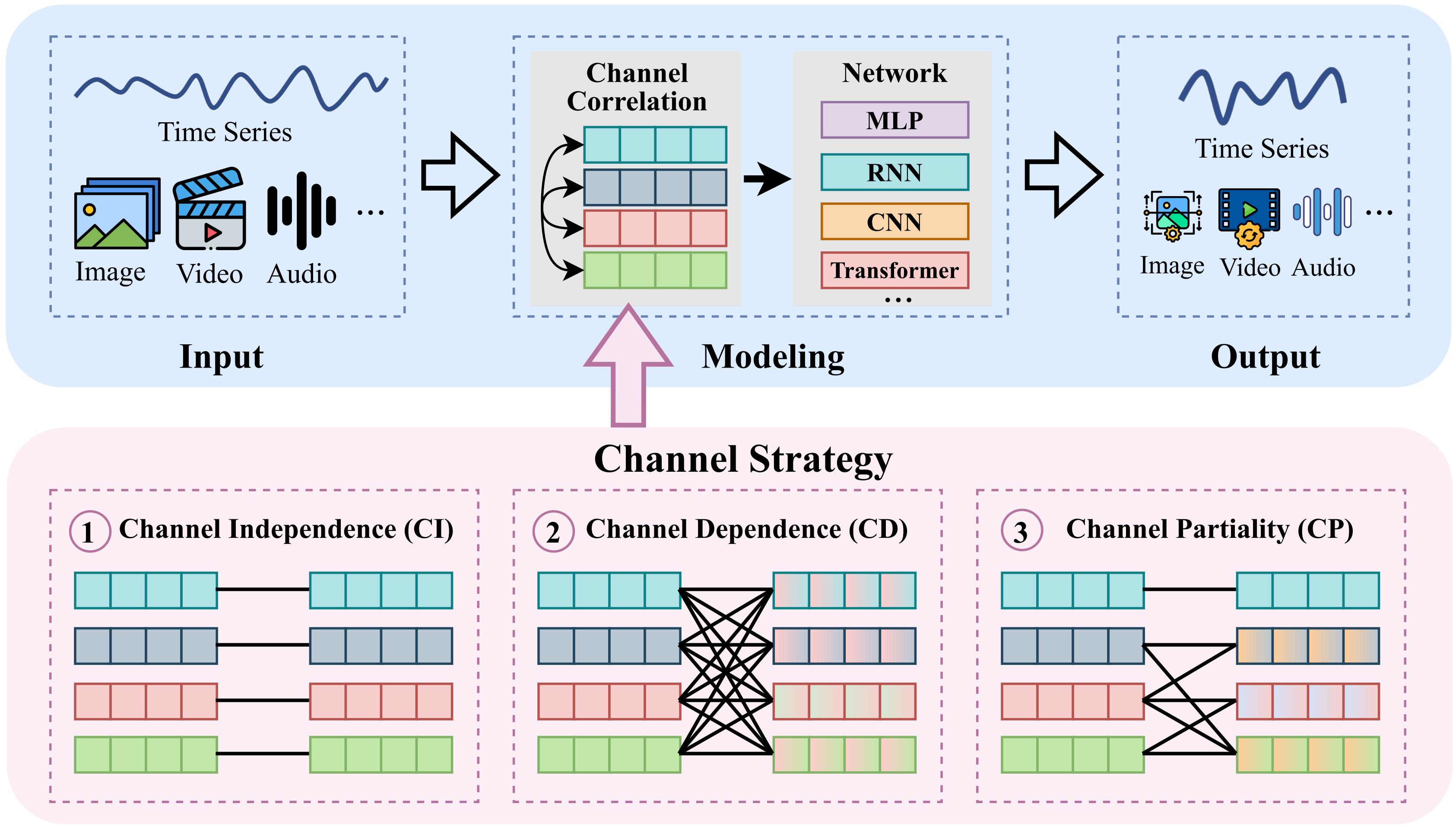}
    \caption{Overview of existing channel strategies.}
    \label{fig:channel_strategy}
\end{figure}

Recently, channel strategies have emerged as a key research direction in multivariate prediction. Leveraging channel correlations can significantly enhance prediction accuracy, especially for specific channels. For instance, in precipitation forecasting, by integrating multi-source data (e.g., temperature, humidity, atmospheric pressure, and wind speed), meteorological models can more accurately identify the occurrence and intensity of rainfall. The interconnections among different factors provide a more comprehensive framework for modeling rainfall dynamics \cite{rasp2020weatherbench}.

The concept of channel strategies has been introduced in multivariate time series forecasting \cite{qiu2025comprehensive, han2024capacity}. As illustrated in Fig.~\ref{fig:channel_strategy}, channel strategies can be classified into three types:
\begin{itemize}
\item  Channel Independence (CI): processes each channel independently without considering potential interactions or correlations; 
\item  Channel Dependence (CD): treats all channels as a unified entity, assuming they are interrelated and mutually dependent; 
\item  Channel Partiality (CP): each channel maintains some degree of independence while being influenced by other related channels. 
\end{itemize}
These strategies are often embedded in DL networks to enhance model performance. For example, SENet \cite{hu2018squeeze} pioneered the channel attention mechanism via the Squeeze-and-Excitation block, effectively modeling channel-wise dependencies to boost the representation ability of CNNs. Similarly, iTransformer \cite{liu2024itransformer} integrates channel information into the Transformer encoder, treating individual time series as tokens and applying self-attention mechanisms to capture inter-variable relationships, thereby improving the generalization ability of Transformers across different variables.

SWH retrieval is also a multivariate prediction task, which essentially investigates the influence of SWH-related variables in SWH prediction. While existing deep learning-based SWH models primarily utilize CYGNSS data, they often overlook inherent channel correlations despite high inter-channel similarity. This paper proposes, for the first time, a four-channel modeling paradigm for SWH retrieval. We compare the characteristics of single-channel and four-channel models, and explore the effects of CI and CD strategies on the performance of the four-channel model.

\section{Methodology} \label{Methodology}
\begin{figure*}[!ht]
    \centering
    \includegraphics[width=\linewidth]{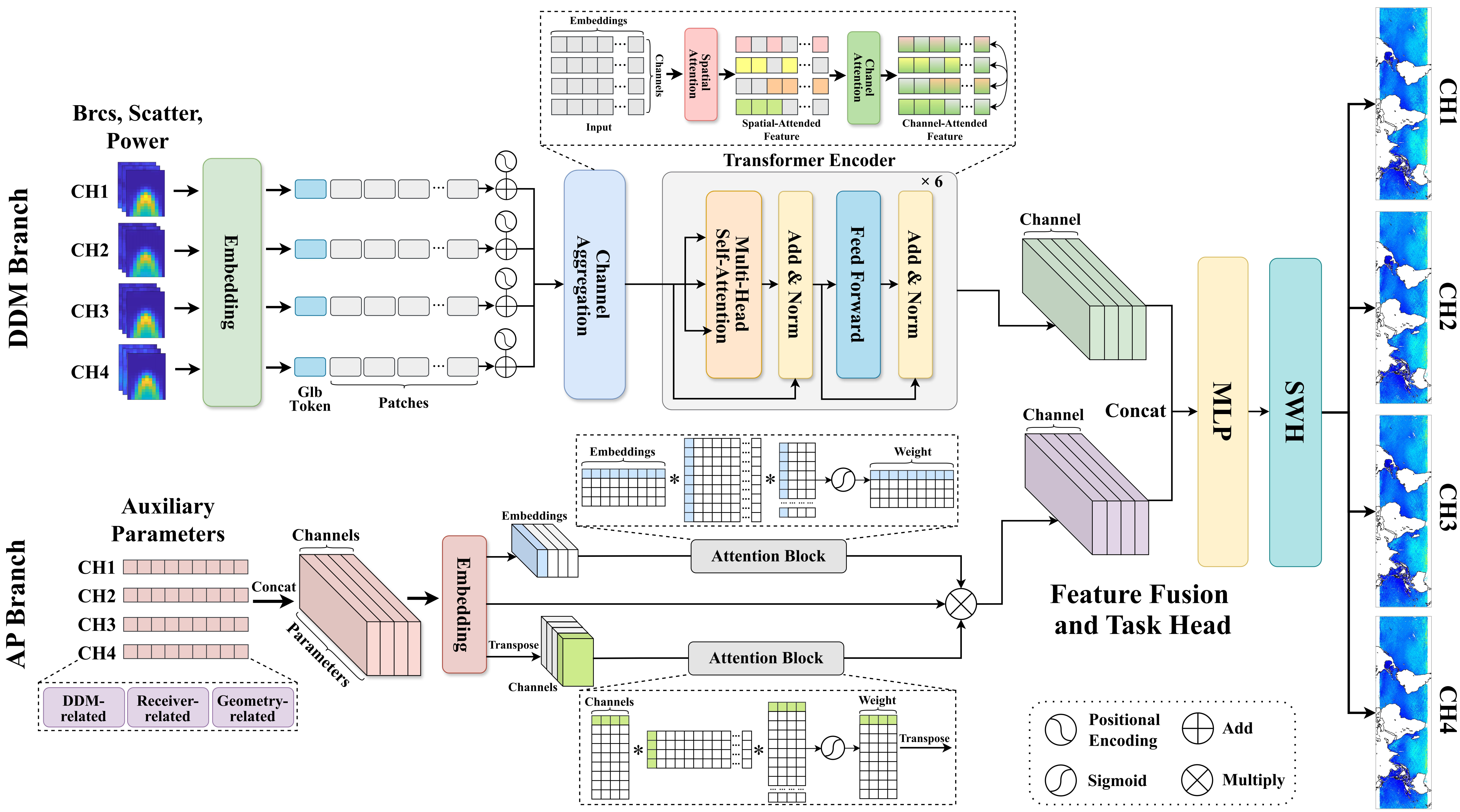}
    \caption{Overall framework of the SCAWaveNet.}
    \label{fig:overall_framework}
\end{figure*}

This section provides a comprehensive overview of the SCAWaveNet architecture, as illustrated in Fig.~\ref{fig:overall_framework}. The model consists of three core components: (1) DDM Branch: employs a Transformer Encoder to apply spatial and channel attention mechanisms sequentially to four-channel DDM embeddings; (2) AP Branch: utilizes a lightweight attention module to compute cross-dimensional attention weights and integrate them into auxiliary parameters; (3) Feature Fusion and Task Head: Fuses multi-branch features to generate four-channel SWH predictions. Each component will be detailed in the following subsections.

\subsection{Transformer-based DDM branch}
This branch is designed to effectively capture the spatial and channel-wise features of DDMs. Previous studies have not sufficiently leveraged the intrinsic features of DDMs. On the one hand, existing approaches typically adopt CNNs as feature extraction modules, which are limited in feature representation ability. On the other hand, these approaches focus only on spatial information while overlooking channel-wise information in DDMs. To address these challenges, we integrate the Spatial-Channel Attention (SCA) mechanism into the DDM branch, which includes three components: an embedding block, a channel aggregation block, and an SCA-based Transformer encoder. These components are elaborated in the subsequent sections.

\subsubsection{Embedding}
Given the DDMs of each channel $x \in \mathbb{R}^{T \times W \times H}$, where $T$ denotes the number of DDM types, i.e., BRCS DDM (Brcs), effective scatter area DDM (Scatter) and analog power DDM (Power). Some models treat different types of DDM as ``channels''\cite{qiao2025hybrid, qiao2024wavetransnet, zhao2023ddm}, but we treat channels as the number of signal paths simultaneously received by the receiver. $W$ and $H$ represent the frequency shift and time delay dimensions of the DDMs, respectively. Following the patch embedding strategy in ViT \cite{dosovitskiy2021an}, the DDMs are divided into non-overlapping patches using a 2D convolutional layer. Automatic padding is applied to handle incomplete regions. This process is formulated as:

\begin{equation}
    x_{p} = \text{Conv2D}(x) = [x^{1}_{p}, x^{2}_{p}, \dots, x^{N}_{p}]
    \label{eq:patch_embedding}
\end{equation}
where $x_{p}\in \mathbb{R}^{p \times N}$ denotes the embedding sequence of each channel, $p \in\{1,2,3\}$ is the number of DDM, $N$ is the number of patches in the sequence.

To enhance the Transformer’s ability to capture global dependencies within the DDMs, a global token $x_{glb}$ is introduced into $x_{p}$. This token aggregates contextual information across the sequence, forming a new sequence $x^{\prime}_{p}$ as:

\begin{equation}
    x^{\prime}_{p} = [x_{glb},x^{1}_{p}, x^{2}_{p}, \dots, x^{N}_{p}]
    \label{eq:glb_token}
\end{equation}
where, $x_{glb}$ is defined as a learnable parameter dynamically updated during training.

Since Transformers lack positional awareness, positional encodings are injected into the embeddings to help the model distinguish features at different locations. In this study, we adopt the original positional encoding in Transformer\cite{vaswani2017attention} and add it to $x^{\prime}_{p}$:

\begin{equation}
    E_p = x_p^{\prime} + {PE(x_p^{\prime})}
    \label{eq:channel_embedding}
\end{equation}
where 
\begin{equation}
\left\{
\begin{aligned}
{PE}_{(pos, 2d)} &= \sin\left(\frac{pos}{10000^{2d/d_{model}}}\right) \\
 {PE}_{(pos, 2d+1)} &= \cos\left(\frac{pos}{10000^{2d/d_{model}}}\right)
\end{aligned}
\right.,
\end{equation}
$pos$ is the position, $d$ is the dimension index, $d_{model}$ is the representation dimension of Transformer. Each even-indexed dimension is encoded with a sine function, while each odd dimension is encoded with a cosine function. $PE(x_p^{\prime})$ denotes the positional encoding for DDMs. $E_p$ denotes the final patch embedding in each channel after incorporating positional information.

\subsubsection{Channel Aggregation}
After embedding the DDMs for each channel, we further integrate the embeddings across all channels to form the input of the Transformer Encoder. First, a flatten operation is applied to each channel's $E_p$ as:

\begin{equation}
    F_{c} = \text{Flatten}(E_{p})_{c} = \left[ x_1^1, \ldots, x_1^N, \ldots, x_3^1, \ldots, x_3^N \right]_{c}
    \label{eq:channel_flatten}
\end{equation}
where $c \in \{1, 2, 3, 4\}$ represents the channel index, $F_{c}$ denotes the flattened vector of all DDMs in the $c$-th channel. Then, $F_{c}$ is stacked along a new channel dimension to form the unified input of the Transformer Encoder as:

\begin{equation}
    E = \text{Stack}(F_1, F_2, F_3, F_4, \text{dim}=\text{ch})
    \label{eq:channel_stack}
\end{equation}
where $E \in \mathbb{R}^{4 \times M}$ is the stacked embedding tensor, $M$ is the length of the flattened embedding vector for each channel, and $\text{ch}$ indicates the channel dimension.

\subsubsection{Transformer Encoder}
The aggregated patch embedding is subsequently fed into six Transformer Encoder layers. As illustrated in Fig.~\ref{fig:overall_framework}, each layer includes three blocks: multi-head self-attention (MSA), residual connection and layer normalization (Add \& Norm) and feedforward network (FFN). Detailed descriptions of these modules are provided below.

\begin{figure}[!ht]
    \centering
    \includegraphics[width=\linewidth]{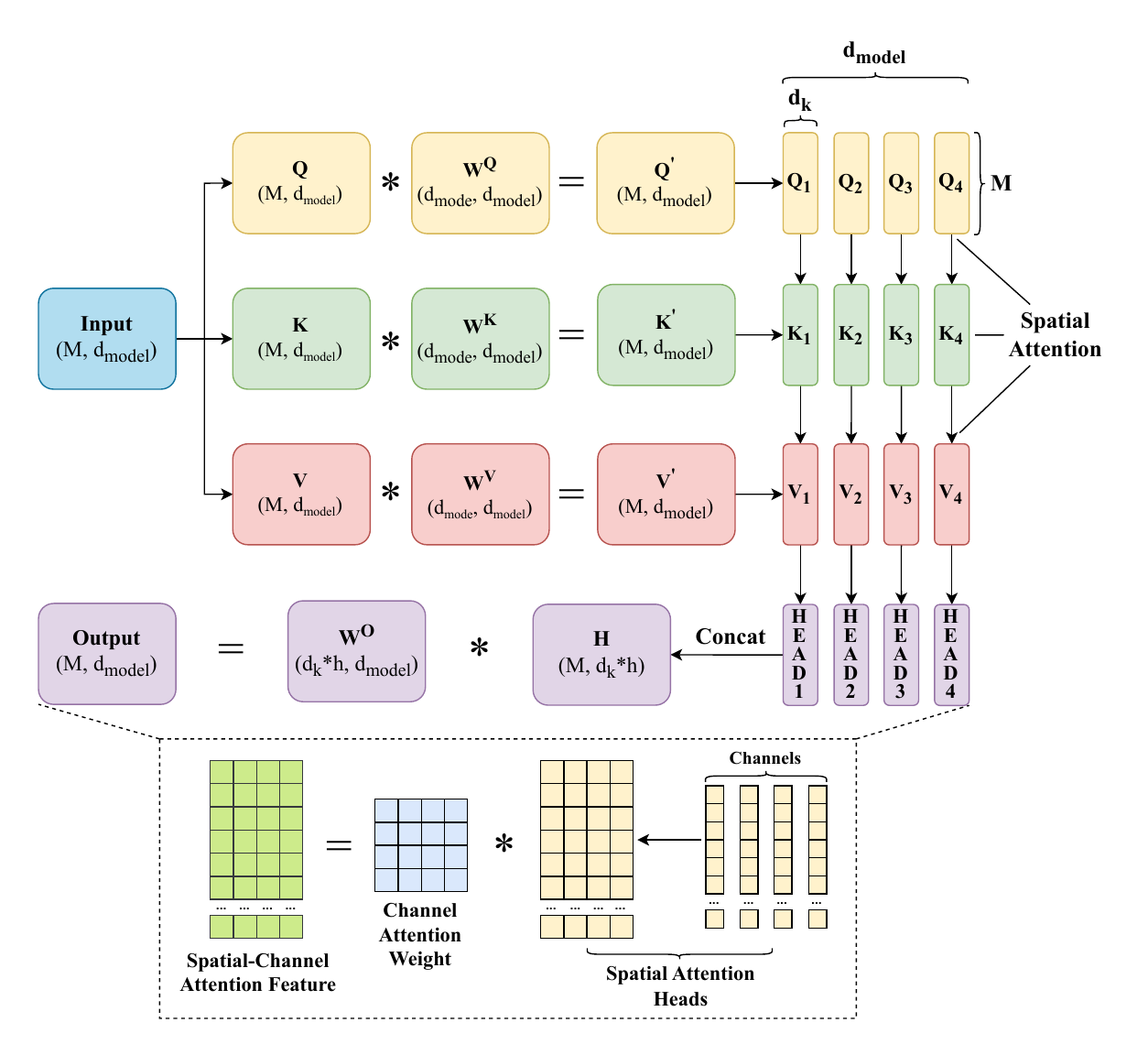}
    \caption{MSA block with spatial-channel attention used in this study.}
    \label{fig:MHA}
\end{figure}

The MSA module is the core component of the Transformer Encoder. Related studies failed to consider channel information in DDM embedding, leading to limited representation of DDMs. To address this, we propose an enhanced MSA block that integrates the SCA mechanism to enable richer feature extraction from DDMs, as shown in Fig.~\ref{fig:MHA}. Specifically, the input embedding $E$ is divided into three equal vectors: query $Q \in \mathbb{R}^{M \times d_{\text{model}}}$, key $K \in \mathbb{R}^{M \times d_{\text{model}}}$, and value $V \in \mathbb{R}^{M \times d_{\text{model}}}$. They are projected by the learnable weight matrix $W^Q$, ${W}^K$, and $W^V$, resulting in transformed representations:

\begin{equation}
\left\{
\begin{aligned}
{Q}^\prime &= {Q} {W}^Q = \left[{Q}_1, {Q}_2, \ldots, {Q}_i\right] \\
{K}^\prime &= {K} {W}^K = \left[{K}_1, {K}_2, \ldots, {K}_i\right] \\
{V}^\prime &= {V} {W}^V = \left[{V}_1, {V}_2, \ldots, {V}_i\right]
\end{aligned}
\right.
\end{equation}
where ${Q}^\prime$, ${K}^\prime$, and ${V}^\prime$ are split along the $d_{\text{model}}$ into $h$ heads. These heads serve as the basic computational units in the MSA block. The $i$-th head ${Q}_i$, ${K}_i$, and ${V}_i$ lie in $\mathbb{R}^{M \times d_k}$, where $d_k = \frac{d_{\text{model}}}{h}$. To integrate SCA into the Transformer Encoder, each channel of the patch embedding is treated as an independent head, with $d_{\text{model}} = 4$ and $h = 4$.

Subsequently, attention scores for the $i$-th channel are computed by ${Q}_i$, ${K}_i$, and ${V}_i$, which essentially applies spatial attention in each channel. The calculation begins with the dot product between ${Q}_i$ and the transpose of ${K}_i$. Then, a \textit{softmax} is applied to obtain the attention weights between ${Q}_i$ and ${K}_i$, which are used to weight ${V}_i$ and produce the final score. When $d_k$ is large, the dot product will become extremely large, causing \textit{softmax} to operate in regions with extremely small gradients. To prevent this phenomenon, the dot product is scaled by a factor of $1/\sqrt{d_k}$~\cite{vaswani2017attention}. The spatial attention computation is defined as:

\begin{equation}
Head_{i} = \mathrm{Softmax}\left( \frac{{Q}_i {K}_i^\top}{\sqrt{d_k}} \right) {V}_i 
\end{equation}
where $Head_{i}$ represents the spatial attention vector for the ${i}$-th channel. All channel outputs are concatenated to form matrix ${H}$, which undergoes linear projection via ${W}^O$ to enable cross-channel information fusion. This projection yields the final MSA output:

\begin{equation}
\mathrm{O}[m,j] = \sum_{i=1}^{4} {H}[m,i] \cdot {W}^O[i,j]
\label{eq:msa_output}
\end{equation}
where $m$ indexes the tokens in the sequence $M$, ${H}[m,i]$ represents the output of the $m$-th token from the $i$-th channel, and ${W}^O[i,j]$ determines the contribution of the $i$-th channel to the $j$-th output dimension. If the weights in ${W}^O$ are non-zero, channel attention is introduced, which affects every element of the output. Therefore, the degree of channel interaction is closely related to ${W}^O$. The projection matrix ${W}^O$ is optimized during backpropagation to dynamically adjust channel interactions. To ensure accurate SWH prediction from all channels, its weights are updated via gradient descent to balance the contributions of each channel to the final output.

The Add \& Norm component consists of two primary operations: residual connection (RC) and layer normalization (LN). For the output of the MSA, LN is first applied to normalize the features of each sample, stabilizing the training process. Then a \textit{dropout} operation is introduced to mitigate overfitting. Finally, an RC is applied by adding the normalized features back to the original MSA output. This procedure is formulated as:

\begin{equation}
{D} = {O} + \mathrm{Dropout}(\mathrm{LN}({O}))
\end{equation}

The output ${D}$ is then fed into an FFN, followed by another Add \& Norm operation, to further refine the feature representation. The FFN consists of two linear layers and a \textit{dropout} operation. Specifically, ${D}$ is first projected through a linear layer, followed by \textit{ReLU} activation and a \textit{dropout} function, and then mapped by a second linear layer to produce the FFN output ${D}_f$. Then, ${D}_f$ is processed by another Add \& Norm component with ${D}$, producing the output of the encoder layer:

\begin{equation}
\left\{
\begin{aligned}
{D}_f &= L_2(\mathrm{Dropout}(\mathrm{ReLU}(L_1({D})))) \\
{D}^\prime &= {D} + \mathrm{Dropout}(\mathrm{LN}({{D}_f}))
\end{aligned}
\right.
\end{equation}
where ${D}^\prime$ is the output of the encoder layer, $L_1 \in \mathbb{R}^{d_{\text{model}} \times d_{\text{ff}}}$ and $L_2 \in \mathbb{R}^{d_{\text{ff}} \times d_{\text{model}}}$ denote the projection matrices. The hidden dimension $d_{\text{ff}}$ is set to 2048, enabling richer feature representations through non-linear transformation and projection.

\begin{table*}[t!]
\caption{Model Input Variables Used in This Study}
\label{tab:input_variables} 
\centering
\renewcommand{\arraystretch}{1.8} 
\setlength{\tabcolsep}{8pt} 
\begin{tabular}{cccc}
\hline
\multirow{2}{*}{\textbf{DDMs}} & \multicolumn{3}{c}{\textbf{Auxiliary Parameters}} \\
\cline{2-4}
& \textbf{DDM-related} & \textbf{Receiver-related} & \textbf{Geometry-related} \\
\hline
brcs, eff\_scatter, power\_analog & ddm\_nbrcs, ddm\_les, ddm\_snr & gps\_eirp, sp\_rx\_gain & sp\_inc\_angle, sp\_lat, sp\_lon, RCG \\
\hline
\end{tabular}
\end{table*}

\subsection{Lightweight Attention-based Auxiliary Parameters Branch}

Previous studies have demonstrated that integrating APs can significantly enhance the accuracy of SWH retrieval \cite{qiao2024wavetransnet,wang2022significant,bu2023combining}. As detailed in Table~\ref{tab:input_variables}, we select nine APs categorized into three types: DDM-related, receiver-related, and geometry-related parameters. The calculation of the range correction gain (RCG) parameter follows the methodology presented in~\cite{bu2023combining,guo2022information}. In contrast to previous research, our APs are acquired from a four-channel DDMI. These parameters are organized into a variable ${A} \in \mathbb{R}^{4 \times 9}$, where each channel contains nine distinct APs. A lightweight attention mechanism is then applied across spatial and channel dimensions to capture inter-channel and spatial dependencies.

As illustrated in Fig.~\ref{fig:overall_framework}, we perform an embedding operation on ${A}$ for each channel to unify the AP tensor in the same space. This process is implemented using a 1D convolutional layer as:

\begin{equation}
    A_{e} = \text{Conv1D}(A) = [a^{1}_{c}, a^{2}_{c}, \dots, a^{9}_{c}]
    \label{eq:patch_embedding}
\end{equation}
where $c$ represents the channel index, $A_{e}$ is the embedding vector, and $a$ is each embedding in each channel. 

Then, $A_{e}$ is split into two equal-sized vectors, denoted as $A_{1}$ and $A_{2}$, which are fed into two distinct attention blocks. Each block includes two projection layers followed by a \textit{sigmoid} activation function, designed to extract attention weights along the spatial and channel dimensions. Specifically, $A_{1}$ undergoes an up-projection to a higher-dimensional space first, then is down-projected back to its original dimension, and finally passes through a \textit{sigmoid} function to generate spatial attention weights. For $A_{2}$, it is first transposed to feed into the attention block, and then transposed back to its original shape after extracting the attention weights. This process is defined as:

\begin{equation}
\left\{
\begin{aligned}
W_{A_{1}} =& \mathrm{Sigmoid}({P}_2({P}_1({A_{1}}))) \\
W_{A_{2}} =& \mathrm{Sigmoid}({P}_4({P}_3({A_{2}})))
\end{aligned}
\right.
\end{equation}
where ${P}_1$, ${P}_2$ are projection matrices applied along the spatial dimension, and ${P}_3$, ${P}_4$ are those applied along the channel dimension. $W_{A_{1}}$ and $W_{A_{2}}$ denote the attention weights computed based on $A_{1}$ and $A_{2}$, respectively. 

Finally, $W_{A_{1}}$ and $W_{A_{2}}$ are applied to the input $A$ via element-wise multiplication. This operation assigns SCA to the four-channel APs, which can be expressed as:
\begin{equation}
{A^\prime} = A \otimes W_{A_{1}} \otimes W_{A_{2}}
\end{equation}
where $\otimes$ denotes element-wise multiplication. The output $A^\prime$ serves as the enhanced APs feature, which integrates both spatial and channel-level characteristics.

\subsection{Feature Fusion and Task Head}

In the previous branches, features from the DDMs and APs were independently extracted to capture spatial and channel-wise information. These two types of features are concatenated along the channel dimension to form a unified representation. This feature representation is then fed into the task head, which is an MLP with 9 hidden layers. The output includes SWH predictions across four channels, expressed as:

\begin{equation}
\hat{y} = [\hat{y}_1, \hat{y}_2, \hat{y}_3, \hat{y}_4] = MLP(\text{Concat}(D^\prime + A^\prime))
\end{equation}
where $\hat{y}$ denotes the predicted SWH, with $\hat{y}_1$ to $\hat{y}_4$ corresponding to the predictions for channels 1 to 4. $D'$ and $A'$ represent the features extracted from the DDMs and APs, respectively.

During training, Huber loss is adopted as the objective function to balance sensitivity to outliers and regression stability. Mathematically, it is defined as:

\begin{equation}
\text{Huber}(y, \hat{y}) = 
\begin{cases} 
\frac{1}{2}(y - \hat{y})^2, & \text{if } |y - \hat{y}| \leq \delta \\
\delta |y - \hat{y}| - \frac{1}{2} \delta^2, & \text{if } |y - \hat{y}| > \delta 
\end{cases}
\end{equation}
where $y$ and $\hat{y}$ denote the ground truth and the predicted values, respectively, $\delta$ is a hyperparameter that controls the transition between the Mean Squared Error (MSE) and Mean Absolute Error (MAE) characteristics. The Huber loss integrates the strengths of MSE and MAE: it behaves quadratically for small for small errors (ensuring smooth gradients) and linearly for large errors (reducing sensitivity to outliers). By tuning $\delta$, the loss function can balance the emphasis on large and small errors. In this study, the training objective is to optimize SWH predictions across all channels. Therefore, the loss is computed using the $\hat{y}$ and $y$ values of the four channels.

\section{Data} \label{Dataset}
\begin{figure}[!ht]
    \centering
    \includegraphics[width=\linewidth]{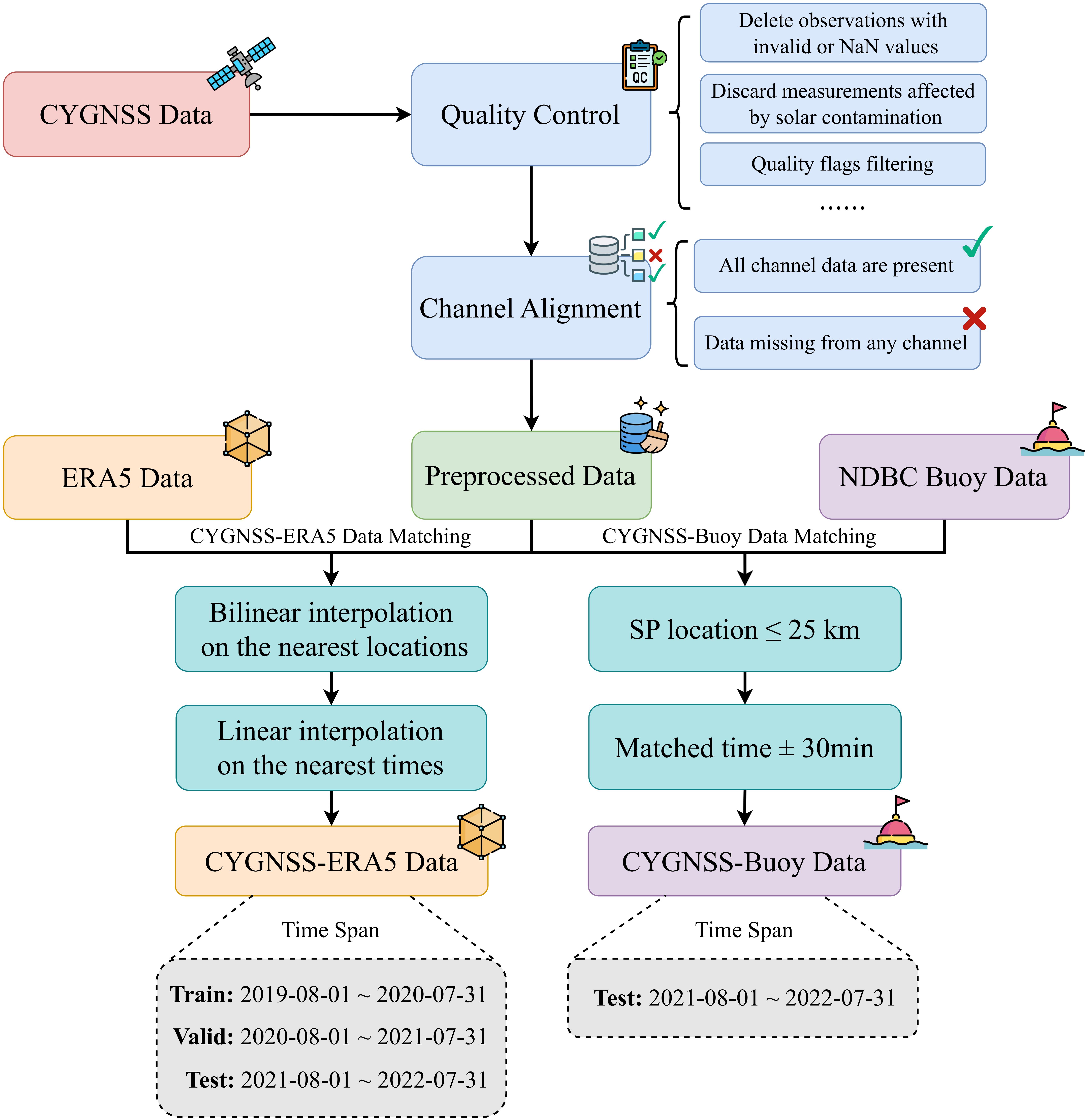}
    \caption{Data processing workflow used in this study.}
    \label{fig:data_process}
\end{figure}

In this section, we provide a comprehensive overview of the dataset employed in this study, including the data source, preprocessing steps, and split strategy. The overall data processing workflow is shown in Fig.~\ref{fig:data_process}.

\subsection{Data Sources}
In this study, GNSS-R observations are obtained from the CYGNSS Level 1 Version 3.2 science data product, provided by the Physical Oceanography Distributed Active Archive Center. CYGNSS, a NASA Earth System Science Pathfinder mission, consists of a constellation of eight microsatellites. These satellites provide near-continuous global coverage through an orbital inclination of approximately $35^\circ$ relative to the equator, with an average revisit time of 7 hours. This inclination enables CYGNSS to measure ocean parameters between nearly $38^\circ$N and $38^\circ$S latitude \cite{CYGNS-L1X32}.

One of the reference datasets used in this study is the ERA5 dataset, the fifth-generation global climate and weather reanalysis product developed by the European Centre for Medium-Range Weather Forecasts\cite{hersbach2023era5}. ERA5 provides three types of SWH products: the combined SWH of wind waves and swell, the SWH of wind waves alone, and the SWH of swell alone. In this study, the combined SWH dataset are employed, which offer a spatial resolution of $0.5^\circ \times 0.5^\circ$ and an hourly temporal resolution. These datasets serve as the foundation for model training, validation, and testing.

In parallel, buoy data obtained from the National Data Buoy Center (NDBC) are used as real-world observations. The NDBC operates 1,350 stations, of which 1,056 have reported in the past 8 hours. We specifically utilized standard meteorological NDBC data, with each station reporting SWH measurements at 30- or 60-minute intervals. Due to the limited availability of buoy data, these measurements are used exclusively as a test set to evaluate the real-world performance of the SWH retrieval model.

\subsection{Data Preprocessing}
After acquiring CYGNSS, ERA5 and NDBC data, a three-step data preprocessing procedure is applied: (1) quality control; (2) channel alignment; (3) CYGNSS-ERA5 and CYGNSS-Buoy data matching.

\subsubsection{Quality Control}
Since ML methods are data-driven and raw CYGNSS observations contain noise that affects the model performance, this study adopts the following quality control criteria to ensure data quality \cite{asgarimehr2022gnss, li2021analysis, clarizia2020statistical}.

\begin{itemize}
    \item Remove all observations containing NaN or Inf values.
    \item Remove observations with fill values (e.g., ``-9999").
    \item Remove observations where any of ddm\_nbrcs, ddm\_les, ddm\_snr or sp\_rx\_gain is less than 0.
    \item Remove observations with RCG values less than 3.
    \item Remove observations affected by solar contamination.
    \item Keep observations where the satellite tracker's attitude state is ``OK''.
    \item Remove observations where the spacecraft's absolute roll exceeds $30^\circ$, yaw exceeds $5^\circ$, or pitch exceeds $10^\circ$.
    \item Remove observations located within 25 km of land to minimize interference from land signals.
    \item Remove observations where any of the first 28 bits in the quality control flag equals to 1.
\end{itemize}

\subsubsection{Channel Alignment}
After data quality control, a channel alignment procedure is applied to synchronize the four-channel CYGNSS observations. For each timestamp, data from all channels are extracted and aligned according to the following rules:

\begin{itemize}
    \item If data from all four channels are present at a given timestamp, they are merged into a synchronized dataset and retained.
    \item If data from any channel are missing at a certain timestamp, the entire observation for that timestamp is discarded.
\end{itemize}

This process ensures that only complete four-channel datasets are retained for subsequent analysis.

\subsubsection{Data Matching}
Following the channel alignment step, we perform reference value matching with the CYGNSS observations. Since CYGNSS data are spatio-temporal, data matching is conducted in both temporal and spatial dimensions. In this study, CYGNSS data are matched with ERA5 and NDBC buoy datasets as follows:

\begin{itemize}
    \item CYGNSS-ERA5 data matching: For each four-channel CYGNSS observation, we use its spatio-temporal information to first identify the two nearest ERA5 data points in both spatial and time. Next, at these points, we apply bilinear interpolation in the spatial domain followed by linear interpolation in the temporal domain to obtain the spatio-temporal reference value.
    
    \item CYGNSS-Buoy data matching: The temporal and spatial matching thresholds for CYGNSS and NDBC buoy data are set to 30 minutes and 25 km, whereby the CYGNSS specular point must be within 25 km of the buoy location and the observation time must fall within ± 30 minutes of the buoy measurement. Due to the stringent requirement of aligning all four CYGNSS channels with identical timestamps, data matching is performed independently for each channel.
\end{itemize}

Due to the fact that the CYGNSS data exhibit signal saturation when SWH \textgreater 8 m, these measurements offer limited modeling value. Therefore, we exclude all data points with SWH values above this threshold.

\subsection{Data Split}
In this study, the temporal coverage of CYGNSS and ERA5 data spans from August 1, 2019, to July 31, 2022. To ensure fairness in model training and to account for potential seasonal variations, the training, validation, and test sets were strictly divided into non-overlapping time periods. Specifically:
\begin{itemize}
    \item The training set covers August 1, 2019, to July 31, 2020;
    \item The validation set spans August 1, 2020, to July 31, 2021;
    \item The test set ranges from August 1, 2021, to July 31, 2022.
\end{itemize}

Given that the dataset contains four-channel data, training a model on the full yearly dataset would be computationally expensive. Therefore, we randomly sampled 3 million, 0.5 million, and 3 million data points to form the final training, validation, and test sets, respectively. Furthermore, to ensure consistency in model evaluation, the buoy data was aligned with the temporal range of the test set.

\begin{figure}[!ht]
    \centering
    \includegraphics[width=\linewidth]{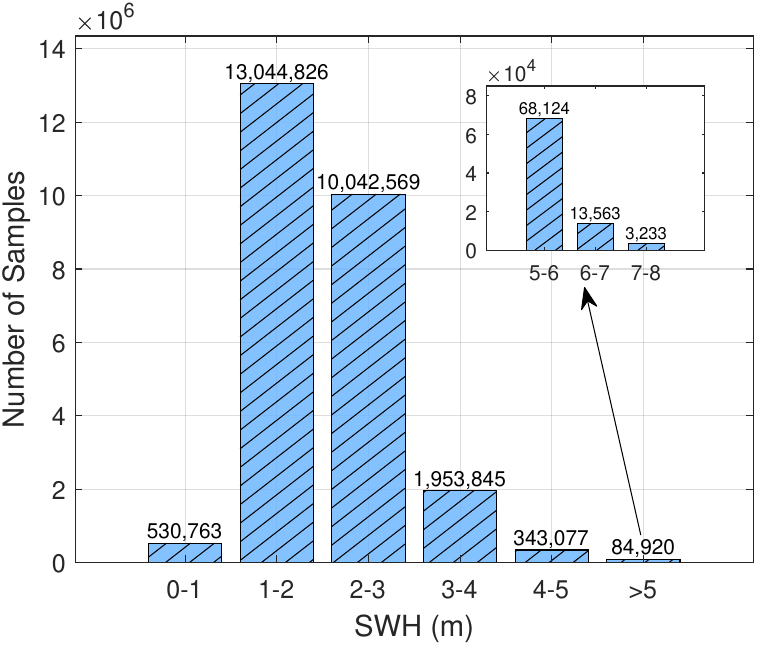}
    \caption{SWH distribution of CYGNSS-ERA5 data.}
    \label{fig:SWH distribution}
\end{figure}

To investigate the distribution of SWH in the GNSS-R data, Fig.~\ref{fig:SWH distribution} presents a histogram of the entire CYGNSS-ERA5 dataset, which contains 6.5 million four-channel data groups (26 million data points). A subplot on the right specifically highlights the distribution of SWH values above 5 m. The histogram shows that the SWH in the data follows a long-tailed distribution, with 88.80\% of the samples concentrated in the 1–3 m range. Samples outside this range are relatively sparse, especially for SWH values above 5 m, which account for only 0.33\% of the CYGNSS-ERA5 dataset.

\section{Experiments} \label{Experiments}
In this section, we present a detailed description of the experimental setup and conduct a comprehensive analysis of the SWH retrieval performance of the proposed SCAWaveNet model and other DL models under various conditions.

\subsection{Experimental Setup}
In the experiments, five metrics are utilized to evaluate the effectiveness of different methods: Root Mean Square Error (RMSE), Mean Absolute Error (MAE), Mean Absolute Percentage Error (MAPE), Bias, and Correlation Coefficient (CC). RMSE, MAE, MAPE, and Bias are applied to assess the error between the predicted and reference SWH from different perspectives:
\begin{itemize}
    \item RMSE reflects the overall deviation;
    \item MAE captures the average magnitude of absolute errors;
    \item MAPE measures the relative error;
    \item Bias indicates the systematic bias in model predictions;
    \item CC quantifies the linear relationship between predictions and references, with higher values indicating a stronger positive correlation.    
\end{itemize}
Lower values of RMSE, MAE, and MAPE indicate a higher prediction accuracy. These metrics are defined as follows: 

\begin{equation}
   \text{RMSE}(\hat{y}, y) = \sqrt{\frac{1}{n} \sum_{i=1}^{n} (\hat{y}_i- y_i)^2}
   \label{eq:rmse}
\end{equation}
\begin{equation}
   \text{MAE}(\hat{y}, y) = \frac{1}{n} \sum_{i=1}^{n} \left|\hat{y}_i- y_i\right|
   \label{eq:mae}
\end{equation}
\begin{equation}
   \text{Bias}(\hat{y}, y) = \frac{1}{n} \sum_{i=1}^{n} (\hat{y}_i- y_i)
   \label{eq:bias}
\end{equation}
\begin{equation}
   \text{MAPE}(\hat{y}, y) = \frac{100\%}{n} \sum_{i=1}^{n} \left| \frac{\hat{y}_i- y_i}{y_i} \right|
   \label{eq:mape}
\end{equation}
\begin{equation}
   CC = \frac{\sum_{i=1}^{n} (\hat{y}_i - \overline{\hat{y}})(y_{i} - \overline{y})}{\sqrt{\sum_{i=1}^{n} (\hat{y}_i - \overline{\hat{y}})^2 \sum_{i=1}^{n} (y_{i}-\overline{y})^2 }}
   \label{eq:mape}
\end{equation}
where $n$ denotes the number of data samples, $\hat{y}_i$ and $y_{i}$ represent the predicted and reference SWH values, respectively. $\overline{\hat{y}}$ and $\overline{y}$ denote the mean values of $\hat{y}_i$ and $y_{i}$, respectively.

Since the modeling paradigm proposed in this study differs from previous DL models, all experimental models are divided into two groups. The first group consists of single-channel models. We adopt CNN-ConvLSTM-FCN \cite{bu2024estimating}, Multi-scale Conv-BiLSTM \cite{wang2024ocean}, WaveTransNet \cite{qiao2024wavetransnet}, ViT-Wave \cite{zhou2024enhancing}, as well as CNN and MLP networks, as baseline models.  The second group consists of four-channel models. Along with the proposed architecture, CNN and MLP are also adopted as backbone networks to extract features from DDMs and APs in both channel and spatial dimensions, serving as baselines.

To distinguish these models in subsequent experiments, we refer to the single-channel models as CNN-ConvLSTM (CNN-ConvLSTM-FCN), Conv-BiLSTM (Multi-scale Conv-BiLSTM), WaveTransNet, ViT-Wave, $\text{MLP}_{s}$ and $\text{CNN}_{s}$. The four-channel models are denoted as $\text{MLP}_{f}$, $\text{CNN}_{f}$ and SCAWaveNet. ${\rm CNN}_s$ and ${\rm MLP}_s$ are designed with the same number of layers and parameters settings as ${\rm CNN}_f$ and ${\rm MLP}_f$, respectively, to facilitate a fair comparison of different modeling approaches. Additionally, we employ two modeling strategies, i.e., Channel Independence (CI) and Channel Dependence (CD), within the four-channel models to explore how inter-channel information interaction influences model performance. The CI strategy extracts features from each channel independently, ignoring inter-channel correlations. The CD strategy treats all channels as a unified entity, allowing full interaction and dependency among channels during modeling.

\begin{table*}[t!]
\caption{Overall performance comparison of all the models}
\label{model_comparison}
\centering
\setlength{\tabcolsep}{3pt} 
\renewcommand{\arraystretch}{1.1}
\begin{tabularx}{\textwidth}{*{15}{>{\centering\arraybackslash}X}}
\toprule
\multirow{2.5}{*}{\textbf{Metric}} & \multirow{2.5}{*}{\textbf{Ch.}} & \multirow{2.5}{*}{\textbf{\shortstack{$\text{MLP}_s$}}} & \multirow{2.5}{*}{\textbf{\shortstack{$\text{CNN}_s$}}} & \multirow{2.5}{*}{\textbf{\shortstack{Conv-\\BiLSTM}}} & \multirow{2.5}{*}{\textbf{\shortstack{CNN-Conv\\LSTM}}} & \multirow{2.5}{*}{\textbf{\shortstack{ViT-\\Wave}}} & \multirow{2.5}{*}{\textbf{\shortstack{Wave\\TransNet}}} & \multicolumn{2}{c}{\multirow{1}{*}{\textbf{\shortstack{$\text{MLP}_f$}}}} & \multicolumn{2}{c}{\multirow{1}{*}{\textbf{\shortstack{$\text{CNN}_f$}}}} & \multicolumn{2}{c}{\multirow{1}{*}{\textbf{\shortstack{SCAWaveNet(Ours)}}}} \\

\cmidrule(lr){9-10} \cmidrule(lr){11-12} \cmidrule(lr){13-14}
& & & & & & & & \multirow{1.1}{*}{CI} & \multirow{1.1}{*}{CD} & \multirow{1.1}{*}{CI} & \multirow{1.1}{*}{CD} & \multirow{1.1}{*}{CI} & \multirow{1.1}{*}{CD} \\

\midrule
\multirow{5}{*}{\shortstack{RMSE(m)$\downarrow$}} 
& Ch1 & 0.481 & 0.470 & 0.467 & 0.461 & 0.463 & 0.453 & 0.483 & 0.470 & 0.470 & 0.454 & 0.452 & \textbf{0.440}\\
& Ch2 & 0.481 & 0.468 & 0.469 & 0.463 & 0.462 & 0.456 & 0.477 & 0.464 & 0.472 & 0.453 & 0.451 & \textbf{0.438}\\
& Ch3 & 0.480 & 0.469 & 0.465 & 0.460 & 0.461 & 0.454 & 0.480 & 0.466 & 0.469 & 0.454 & 0.449 & \textbf{0.437}\\
& Ch4 & 0.478 & 0.472 & 0.463 & 0.458 & 0.461 & 0.452 & 0.472 & 0.461 & 0.466 & 0.451 & 0.447 & \textbf{0.436}\\
& Avg. & 0.480 & 0.470 & 0.466 & 0.460 & 0.462 & 0.454 & 0.478 & 0.465 & 0.469 & 0.453 & 0.450 & \textbf{0.438}\\

\midrule
\multirow{5}{*}{\shortstack{MAE(m)$\downarrow$}} 
& Ch1 & 0.370 & 0.360 & 0.365 & 0.351 & 0.356 & 0.352 & 0.366 & 0.363 & 0.357 & 0.355 & 0.345 & \textbf{0.333}\\
& Ch2 & 0.368 & 0.358 & 0.366 & 0.353 & 0.354 & 0.355 & 0.364 & 0.357 & 0.359 & 0.351 & 0.343 & \textbf{0.330}\\
& Ch3 & 0.366 & 0.359 & 0.362 & 0.349 & 0.353 & 0.353 & 0.366 & 0.359 & 0.360 & 0.352 & 0.339 & \textbf{0.329}\\
& Ch4 & 0.367 & 0.360 & 0.357 & 0.348 & 0.352 & 0.350 & 0.362 & 0.353 & 0.355 & 0.350 & 0.338 & \textbf{0.328}\\
& Avg. & 0.368 & 0.359 & 0.363 & 0.350 & 0.354 & 0.353 & 0.365 & 0.358 & 0.358 & 0.352 & 0.341 & \textbf{0.330}\\

\midrule
\multirow{5}{*}{\shortstack{Bias(m)}} 
& Ch1 & -0.101 & -0.091 & 0.068 & 0.044 & -0.087 & 0.019 & -0.117 & -0.069 & -0.093 & 0.031 & 0.021 & \textbf{0.019}\\
& Ch2 & -0.096 & -0.085 & 0.071 & 0.055 & -0.065 & 0.039 & -0.106 & -0.049 & -0.084 & 0.028 & -0.024 & \textbf{-0.017}\\
& Ch3 & -0.090 & -0.087 & 0.064 & 0.041 & -0.062 & 0.019 & -0.096 & -0.062 & -0.082 & 0.022 & -0.019 & \textbf{-0.016}\\
& Ch4 & -0.085 & -0.092 & 0.040 & 0.023 & -0.063 & 0.007 & -0.089 & -0.045 & -0.070 & 0.015 & -0.018 & \textbf{-0.010}\\
& Avg. & -0.093 & -0.089 & 0.061 & 0.041 & -0.070 & 0.021 & -0.102 & -0.056 & -0.082 & 0.024 & -0.017 & \textbf{-0.012}\\
\midrule
\multirow{5}{*}{\shortstack{CC$\uparrow$}} 
& Ch1 & 0.652 & 0.662 & 0.673 & 0.682 & 0.678 & 0.689 & 0.647 & 0.674 & 0.662 & 0.688 & 0.695 & \textbf{0.713}\\
& Ch2 & 0.653 & 0.664 & 0.671 & 0.680 & 0.679 & 0.686 & 0.654 & 0.677 & 0.660 & 0.693 & 0.698 & \textbf{0.715}\\
& Ch3 & 0.654 & 0.660 & 0.673 & 0.683 & 0.681 & 0.688 & 0.650 & 0.678 & 0.665 & 0.690 & 0.700 & \textbf{0.717}\\
& Ch4 & 0.656 & 0.658 & 0.677 & 0.684 & 0.682 & 0.692 & 0.658 & 0.680 & 0.666 & 0.695 & 0.703 & \textbf{0.719}\\
& Avg. & 0.654 & 0.661 & 0.674 & 0.682 & 0.680 & 0.689 & 0.652 & 0.677 & 0.663 & 0.692 & 0.699 & \textbf{0.716}\\
\midrule
\multirow{5}{*}{\shortstack{MAPE(\%)$\downarrow$}} 
& Ch1 & 18.963 & 18.272 & 18.069 & 17.526 & 17.988 & 17.354 & 18.809 & 18.085 & 18.241 & 17.458 & 17.491 & \textbf{16.509}\\
& Ch2 & 18.857 & 18.154 & 18.291 & 17.894 & 17.769 & 17.715 & 18.869 & 18.026 & 18.329 & 17.321 & 17.156 & \textbf{16.396}\\
& Ch3 & 18.791 & 18.236 & 17.892 & 17.503 & 17.692 & 17.525 & 18.625 & 17.946 & 18.185 & 17.234 & 17.052 & \textbf{16.158}\\
& Ch4 & 18.682 & 18.453 & 17.594 & 17.430 & 17.438 & 17.029 & 18.423 & 17.487 & 18.027 & 17.015 & 16.998 & \textbf{15.932}\\
& Avg. & 18.823 & 18.279 & 17.962 & 17.567 & 17.722 & 17.406 & 18.682 & 17.886 & 18.196 & 17.257 & 17.174 & \textbf{16.274}\\
\bottomrule
\end{tabularx}
\end{table*}

In the experiments, we employed the PyTorch framework to train, validate, and test four-channel models. The optimizer is AdamW with a weight decay of $10^{-5}$. During training, the batch size is set to 512, and the maximum number of epochs is 75. To avoid model overfitting and enhance computational efficiency, an early stopping strategy with patience of 15 epochs is applied. The best model checkpoint is determined by the lowest average RMSE across four channels. The Huber loss is adopted with a hyperparameter $\delta = 2.0$. Learning rates were adjusted according to the model architectures: $1.4 \times 10^{-4}$ for SCAWaveNet and $10^{-4}$ for $\text{MLP}_{f}$ and $\text{CNN}_{f}$ (as their significant performance degrades when the learning rate is set to $1.4 \times 10^{-4}$). For single-channel models, we maintain the same deep learning frameworks and experimental settings as those used in the original studies to ensure methodological consistency. The four single-channel models were trained separately using data from each of the four channels. All models were developed on a single NVIDIA GeForce RTX 4080 GPU.

\subsection{ERA5 Data Evaluation}
Based on the ERA5 dataset, both comparative experiments and ablation experiments are provided in detail. To evaluate the algorithm's performance, we also analyzed the number of parameters in the model. 

\subsubsection{Quantitative Experiments}
We first present the quantitative results of all the models on the CYGNSS-ERA5 test set, as shown in Table~\ref{model_comparison}. SCAWaveNet-CD achieves the best results with the lowest average RMSE (0.438 m) across four channels. Compared to single-channel models, it reduces RMSE by 8.75\% relative to $\text{MLP}_{s}$ (0.480 m), 6.81\% relative to $\text{CNN}_{s}$ (0.470 m), 6.01\% relative to Conv-BiLSTM (0.466 m), 4.78\% relative to CNN-ConvLSTM (0.460 m), 5.19\% relative to ViT-Wave (0.462 m), and 3.52\% relative to WaveTransNet (0.454 m). Compared with other four-channel models, it achieves 6.16\% and 3.31\% lower RMSE than $\text{MLP}_{f}$-CD (0.465 m) and $\text{CNN}_{f}$-CD (0.453 m), respectively. Furthermore, it attains the lowest average MAE (0.330 m) and MAPE (16.274\%), indicating superior SWH retrieval accuracy, and it also achieves the highest average CC (0.716), reflecting strong consistency with ERA5 observations. Among single-channel models, WaveTransNet exhibits the best average performance, highlighting the advantage of the Transformer Encoder over CNNs and MLPs in dynamic feature learning and long-range dependency modeling. Despite ViT-Wave employs a Vision Transformer architecture in the DDM branch, it directly concatenates APs with DDM features without feature extraction, resulting in performance degradation. This confirms the importance of effective feature extraction from the auxiliary parameters.

\begin{figure*}[!t]
    \centering    

    \includegraphics[width=0.92\textwidth]{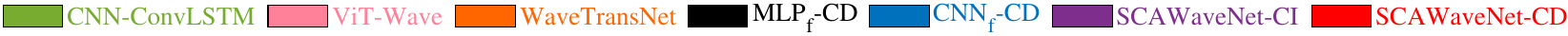}
    \vspace{2mm}

    \centering    
    \begin{minipage}[t]{0.492\textwidth}
        \centering
        \includegraphics[width=\textwidth]{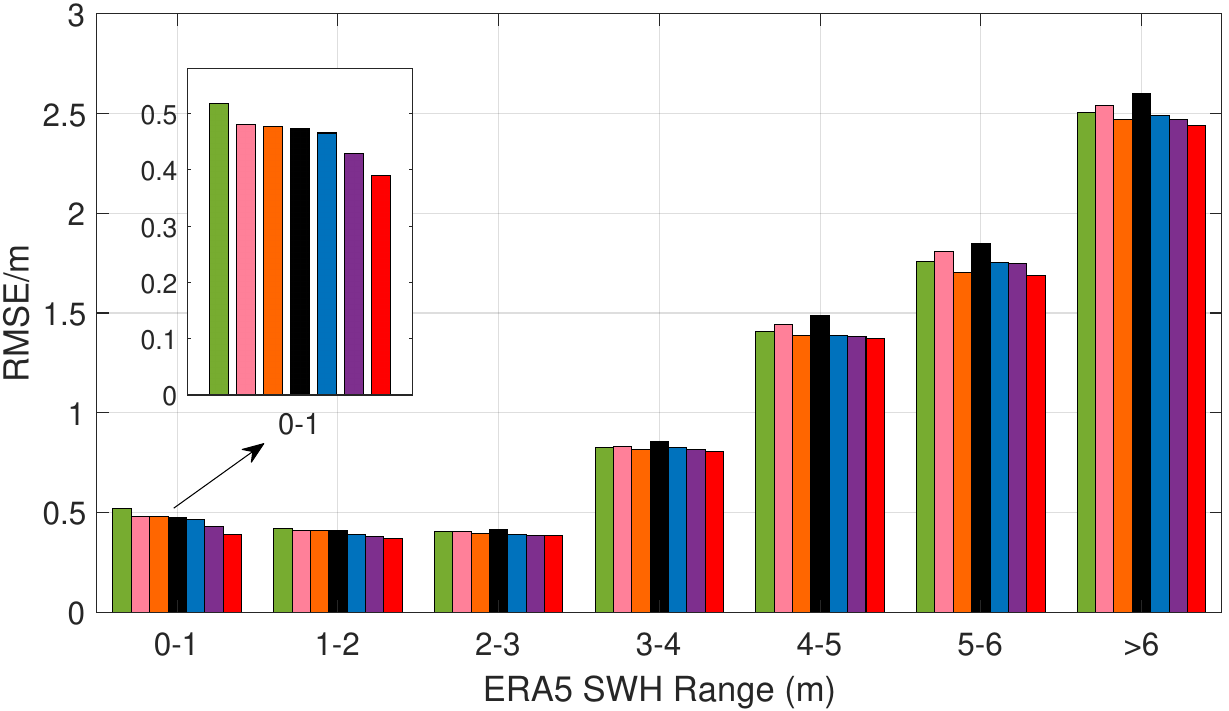}
        \label{fig:diff_range_a}
        \centerline{(a)}
    \end{minipage}
    \hfill
    \begin{minipage}[t]{0.492\textwidth}
        \centering
        \includegraphics[width=\textwidth]{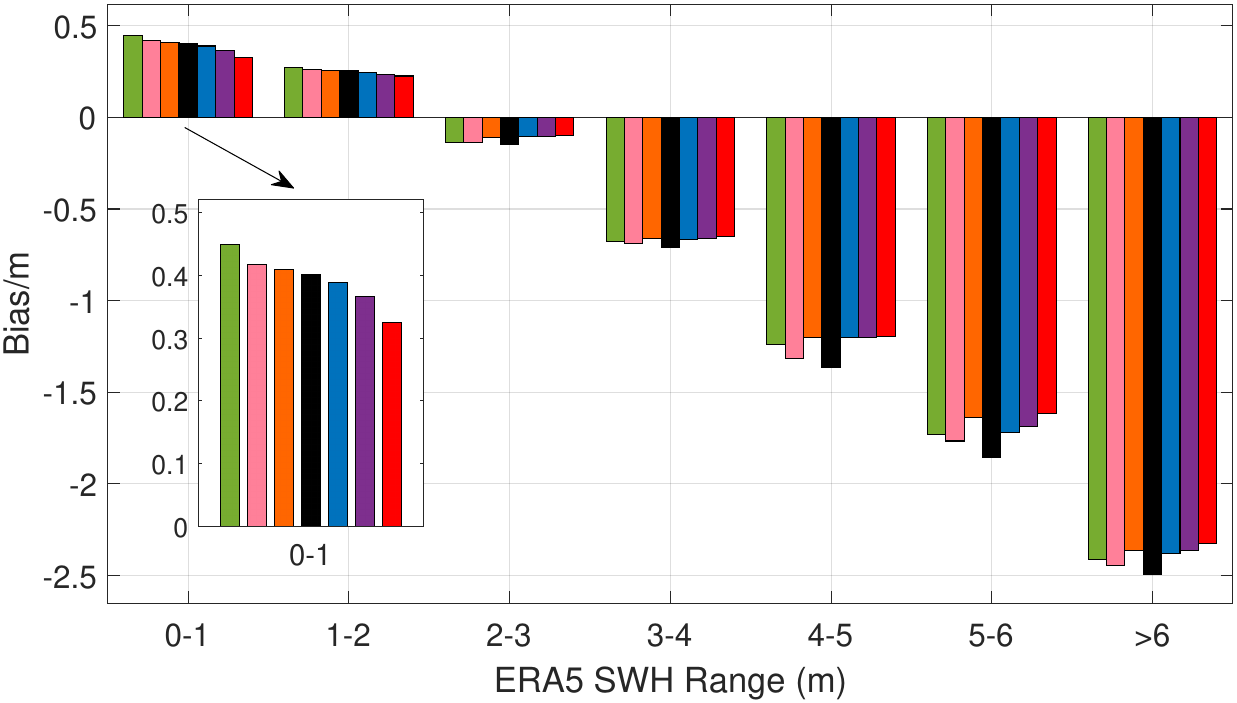}
        \label{fig:diff_range_b}
        \centerline{(b)}
    \end{minipage}

    \caption{Average RMSE and bias for seven models across different ERA5 SWH ranges: (a) RMSE, (b) bias.}
    \label{fig:diff_range}

\end{figure*}

\begin{figure*}[!t]
    \centering
    \begin{minipage}[t]{0.24\textwidth}
        \centering
        \includegraphics[width=\linewidth]{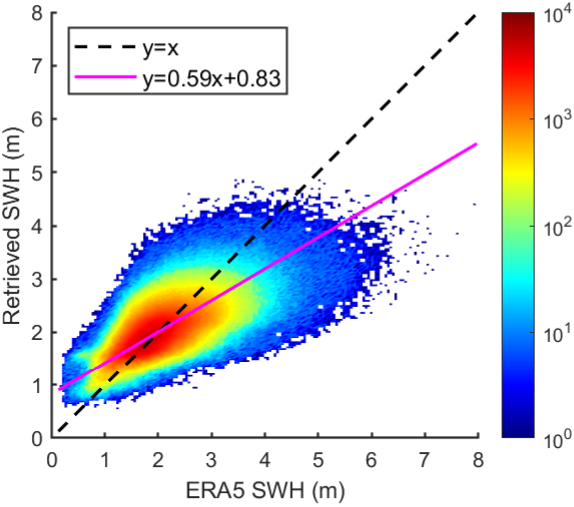}
        \centerline{(a)}
        \label{subfig:ConvLSTM}
    \end{minipage}
    \hfill
    \begin{minipage}[t]{0.24\textwidth}
        \centering
        \includegraphics[width=\linewidth]{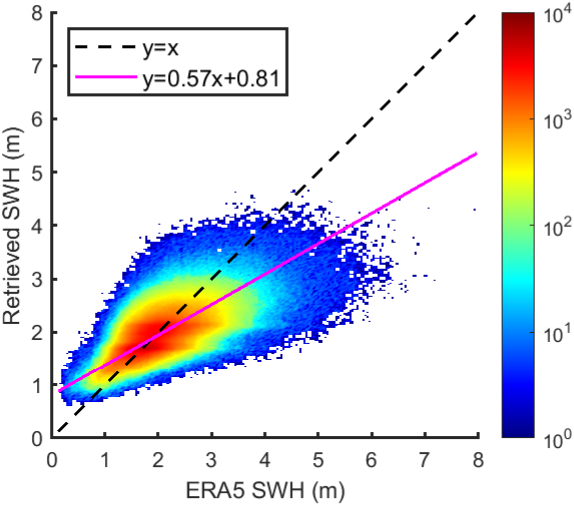}
        \centerline{(b)}
        \label{subfig:vit-wave}
    \end{minipage}
    \hfill
    \begin{minipage}[t]{0.24\textwidth}
        \centering
        \includegraphics[width=\linewidth]{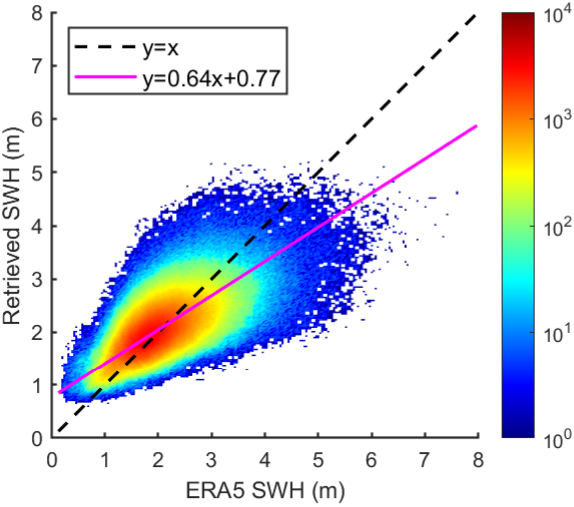}
        \centerline{(c)}
        \label{subfig:wavetransnet}
    \end{minipage}
    \hfill
    \begin{minipage}[t]{0.24\textwidth}
        \centering
        \includegraphics[width=\linewidth]{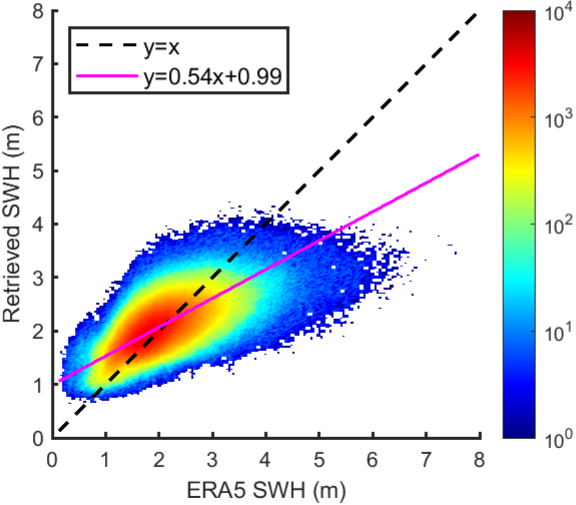}
        \centerline{(d)}
        \label{subfig:MLP}
    \end{minipage}

    \vspace{0mm} 

    \begin{minipage}[t]{0.24\textwidth}
        \centering
        \includegraphics[width=\linewidth]{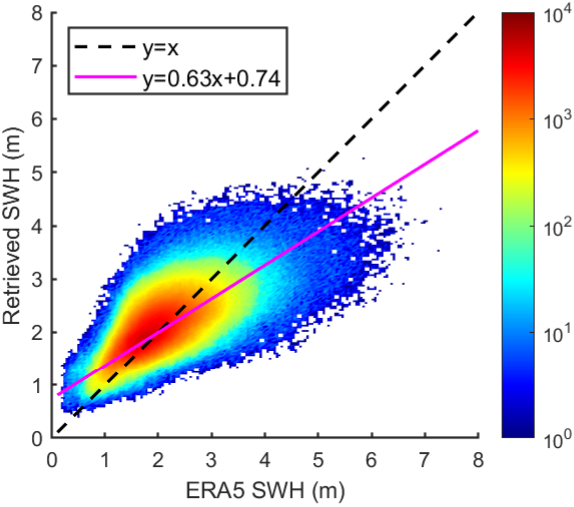}
        \centerline{(e)}
        \label{subfig:CNN}
    \end{minipage}
    \hfill
    \begin{minipage}[t]{0.24\textwidth}
        \centering
        \includegraphics[width=\linewidth]{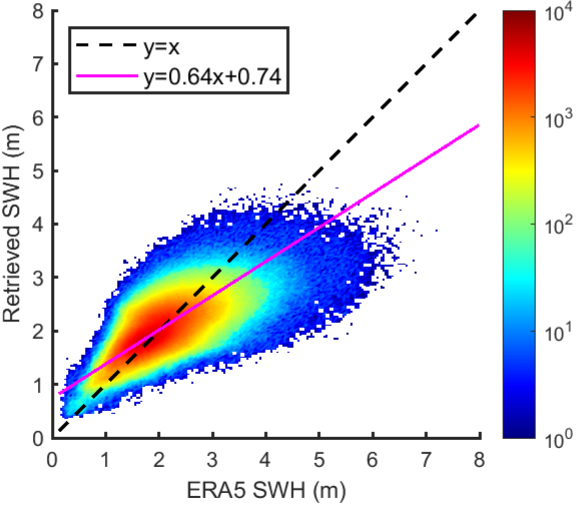}
        \centerline{(f)}
        \label{subfig:SCAWaveNet-CI}
    \end{minipage}
    \hfill
    \begin{minipage}[t]{0.24\textwidth}
        \centering
        \includegraphics[width=\linewidth]{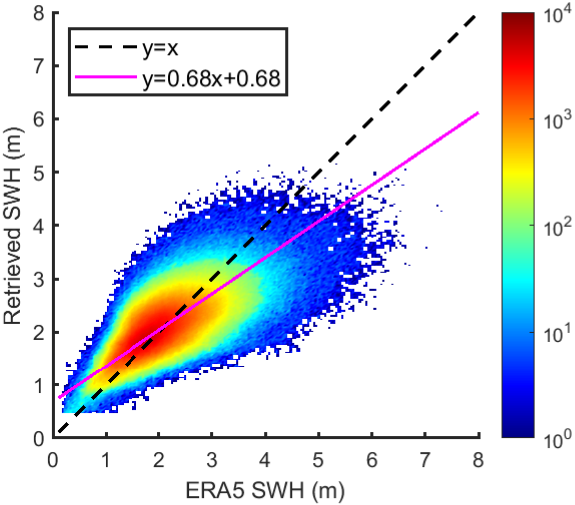}
        \centerline{(g)}
        \label{subfig:SCAWaveNet}
    \end{minipage}
    \hfill
    \begin{minipage}[t]{0.24\textwidth}
        \centering
        \includegraphics[width=\linewidth]{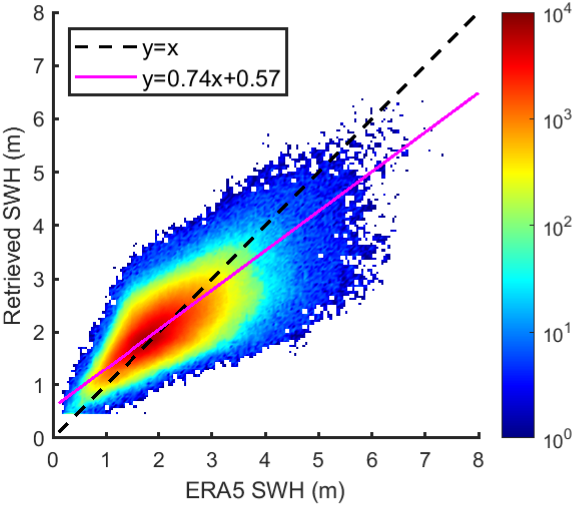}
        \centerline{(h)}
        \label{subfig:SCAWaveNet_wind}
    \end{minipage}

    \caption{Density scatter plots of model predictions for the CYGNSS-ERA5 channel 1 test set: (a) CNN-ConvLSTM, (b) ViT-Wave, (c) WaveTransNet, (d) $\text{MLP}_f$-CD, (e) $\text{CNN}_f$-CD, (f) SCAWaveNet-CI, (g) SCAWaveNet-CD, (h) SCAWaveNet-CD-wind.}
    \label{fig:testset_scatter}
\end{figure*}

From the perspective of modeling strategies, four-channel models using the CD strategy outperform those using the CI strategy. Specifically, SCAWaveNet-CD reduces the average RMSE by 2.67\% compared to SCAWaveNet-CI (0.450 m), $\text{MLP}_{f}$-CD by 3.41\% compared to $\text{MLP}_{f}$-CI (0.478 m), and $\text{CNN}_{f}$-CD by 2.72\% compared to $\text{CNN}_{f}$-CI (0.469 m). These results demonstrate the critical role of channel interaction in improving model performance. Furthermore, $\text{MLP}_{s}$ and $\text{MLP}_{f}$-CI, as well as $\text{CNN}_{s}$ and $\text{CNN}_{f}$-CI, exhibit similar performance. This indicates that four-channel models with the CI strategy behave similarly to the same backbone in single-channel models, with minor differences attributed to variations in modeling objectives.

To analyze in more detail, we compare the prediction performance of selected models across varying SWH conditions. Based on the results in Table~\ref{model_comparison}, we select seven high-performing models for subsequent experiments. Specifically, CNN-ConvLSTM, ViT-Wave, and WaveTransNet are selected from the single-channel models, while $\text{MLP}_f$-CD, $\text{CNN}_f$-CD, SCAWaveNet-CI, and SCAWaveNet-CD are selected from the four-channel models. Fig.~\ref{fig:diff_range} presents histograms of the average RMSE and Bias values for the selected models across four channels within different ERA5 SWH ranges. RMSE and Bias exhibit distinct variation patterns: RMSE values remain relatively low within 0-3 m, but increase progressively when SWH \textgreater 3 m. In contrast, models tend to overestimate SWH in the 0-2 m range, and this overestimation tendency weakens as the SWH increases. Once the SWH \textgreater 2 m, underestimation begins to dominate and becomes increasingly pronounced at higher SWH values.

As illustrated in Fig.~\ref{fig:diff_range}(a), compared to other models, SCAWaveNet-CD achieves the lowest average RMSE across the entire SWH range. Especially in the 0–1 m range, it demonstrates a significant improvement, with an RMSE of 0.389 m, which is 9.34\% lower than SCAWaveNet-CI, 16.35\% lower than $\text{CNN}_f$-CD, and 18.49\% lower than WaveTransNet. A similar pattern is observed in Bias, as shown in Fig.~\ref{fig:diff_range}(b). SCAWaveNet-CD attains a Bias of 0.325 m in the 0-1 m range, outperforming SCAWaveNet-CI by 11.31\%, $\text{CNN}_f$-CD by 16.60\%, and WaveTransNet by 20.56\%. These results demonstrate that the SCA mechanism enhances overall model performance, particularly in the 0-1 m range.

\subsubsection{Qualitative Experiments}
\begin{figure*}[!t]
    \centering    

    \begin{minipage}[t]{\textwidth}
        \centering
        \includegraphics[width=0.6\textwidth]{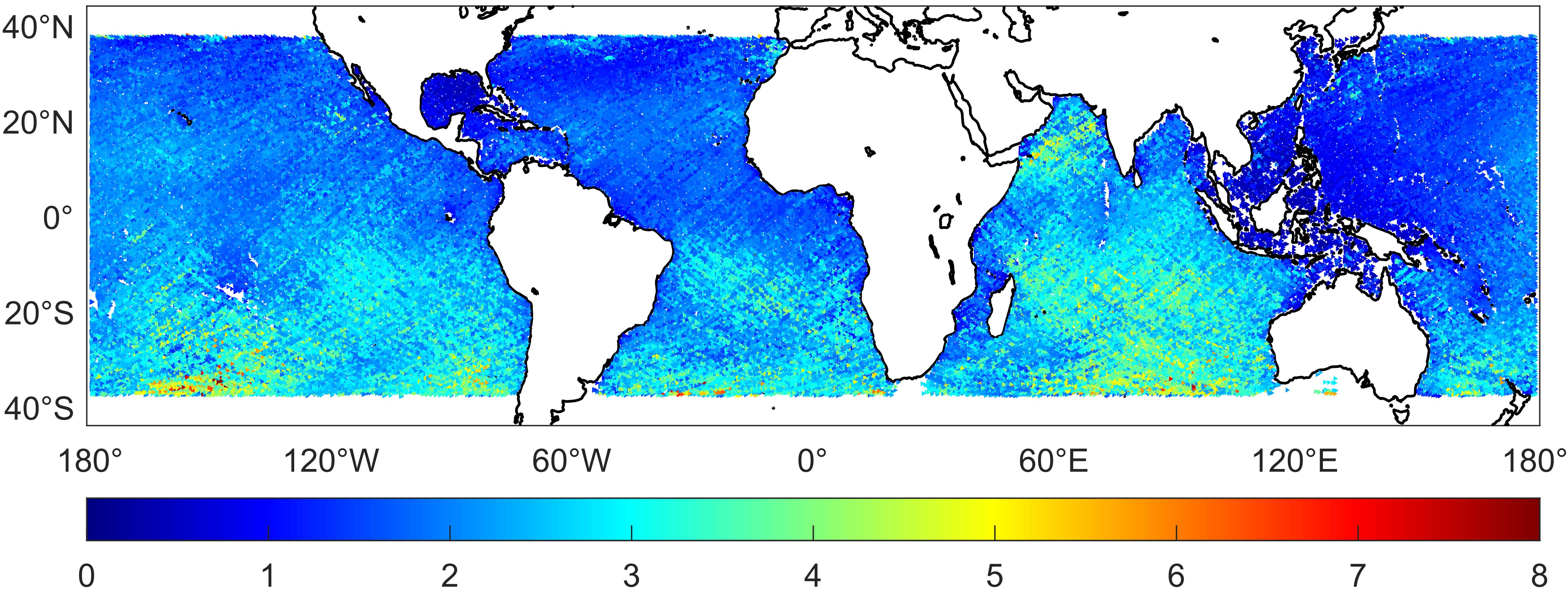} 
        \centerline{(a)}
    \end{minipage}

    \vspace{0.5mm} 

    \begin{minipage}[t]{0.492\textwidth}
        \centering
        \includegraphics[width=\textwidth]{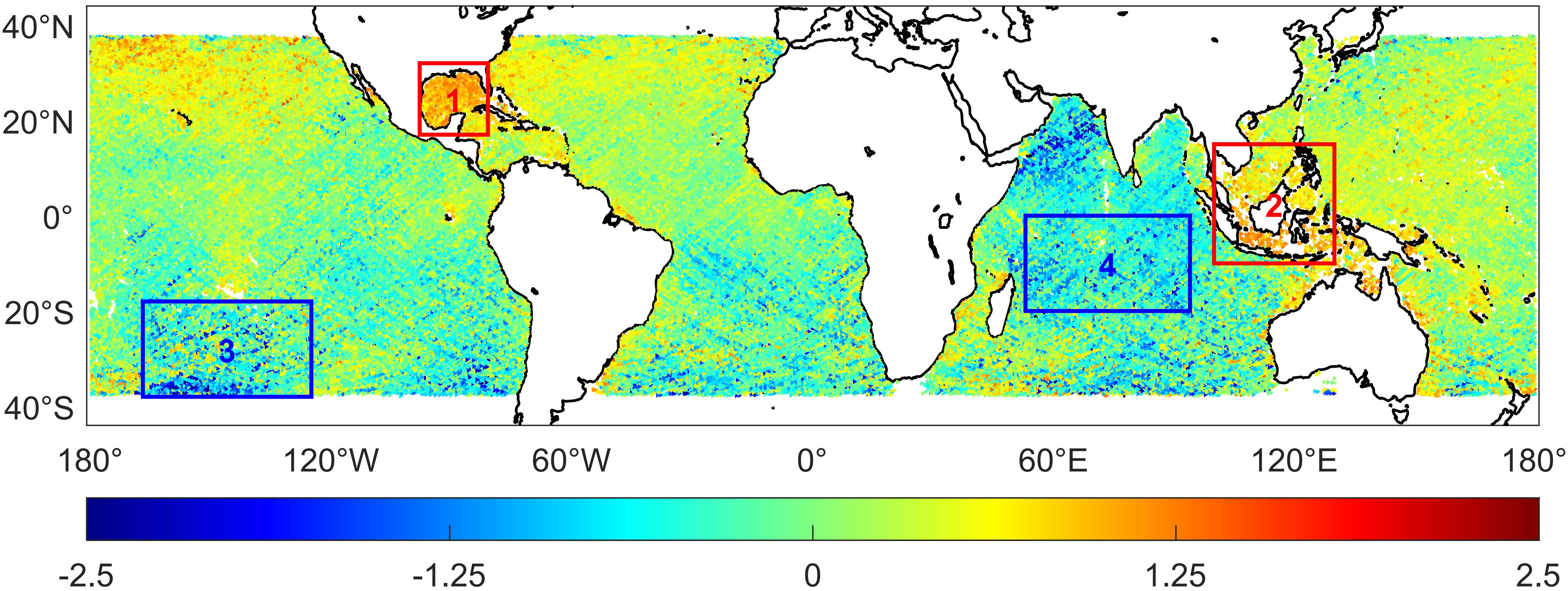}
        \centerline{(b)}
    \end{minipage}
    \hfill
    \begin{minipage}[t]{0.492\textwidth}
        \centering
        \includegraphics[width=\textwidth]{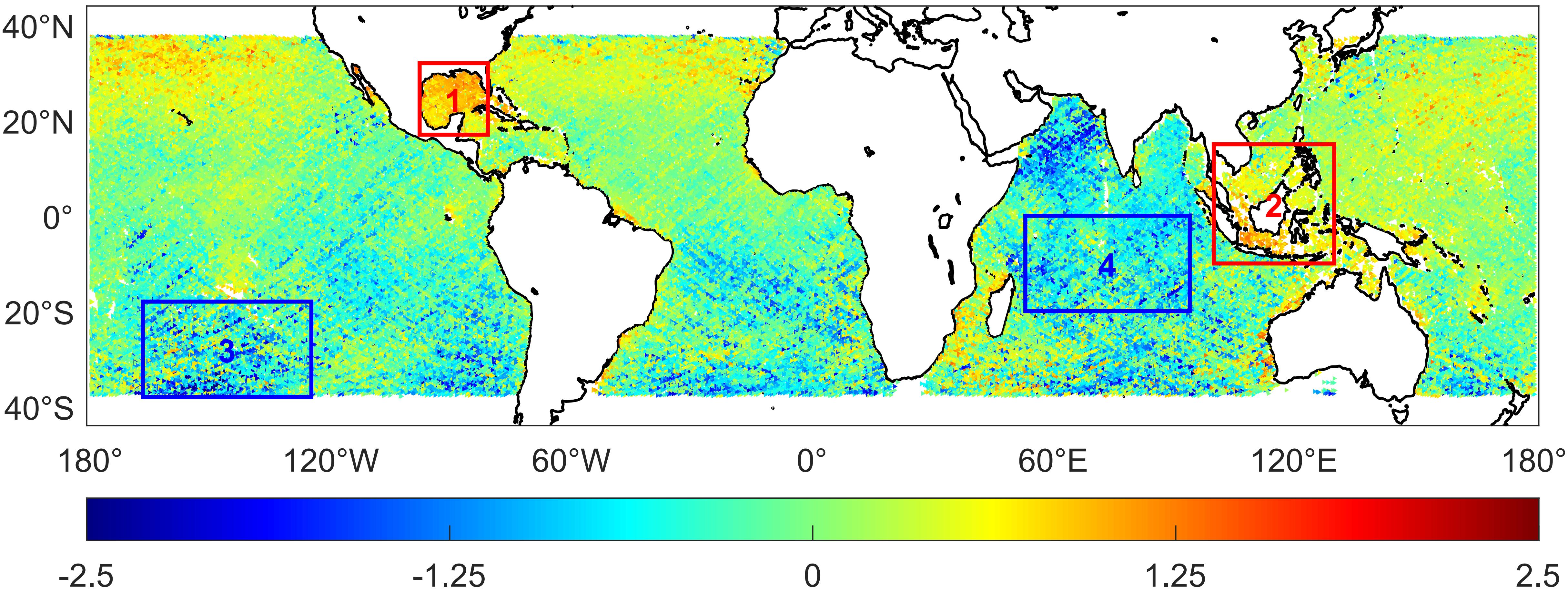}
        \centerline{(c)}
    \end{minipage}

    \vspace{0.5mm} 

    \begin{minipage}[t]{0.492\textwidth}
        \centering
        \includegraphics[width=\textwidth]{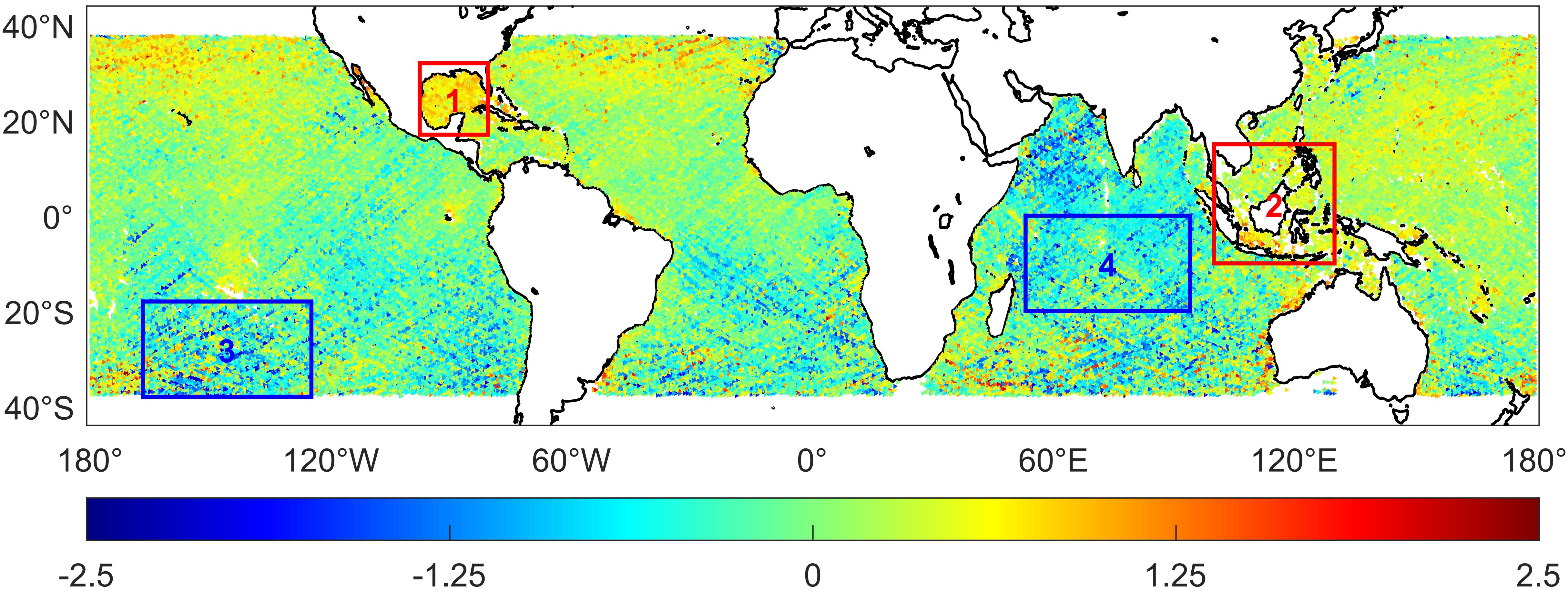}
        \centerline{(d)}
    \end{minipage}
    \hfill
    \begin{minipage}[t]{0.492\textwidth}
        \centering
        \includegraphics[width=\textwidth]{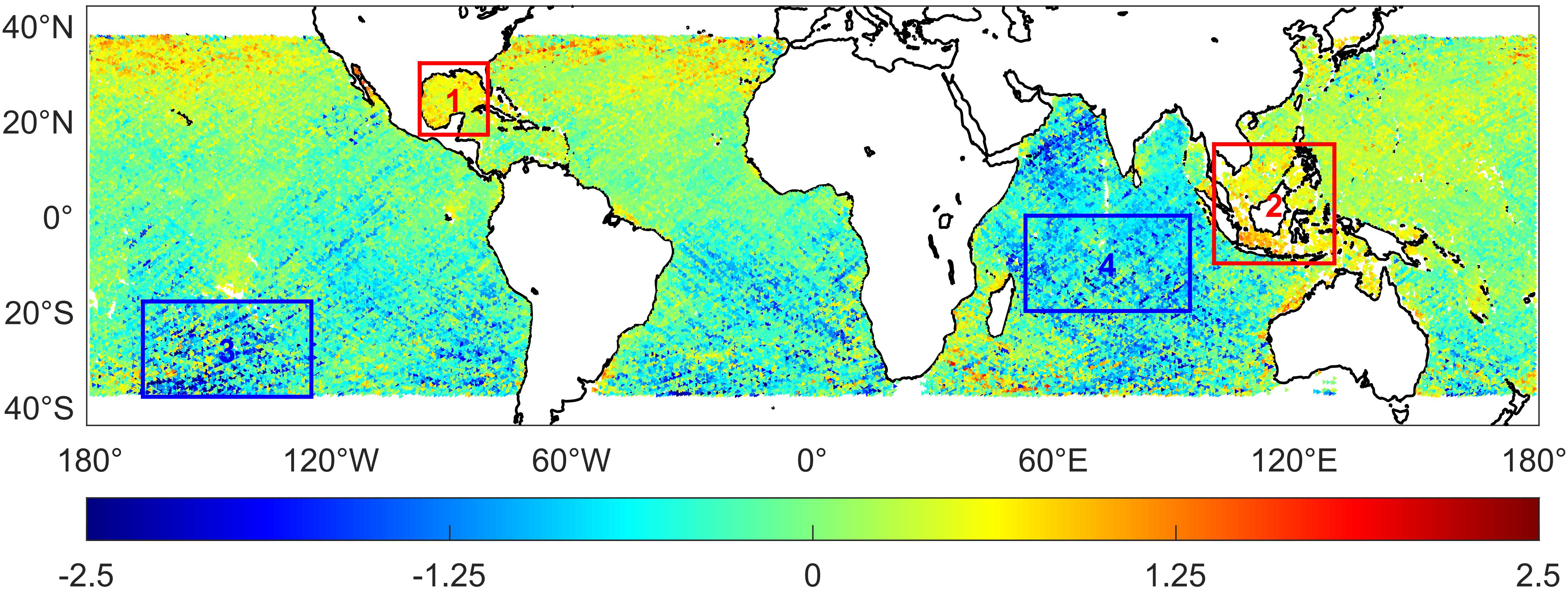}
        \centerline{(e)}
    \end{minipage}

    \vspace{0.5mm} 

    \begin{minipage}[t]{0.492\textwidth}
        \centering
        \includegraphics[width=\textwidth]{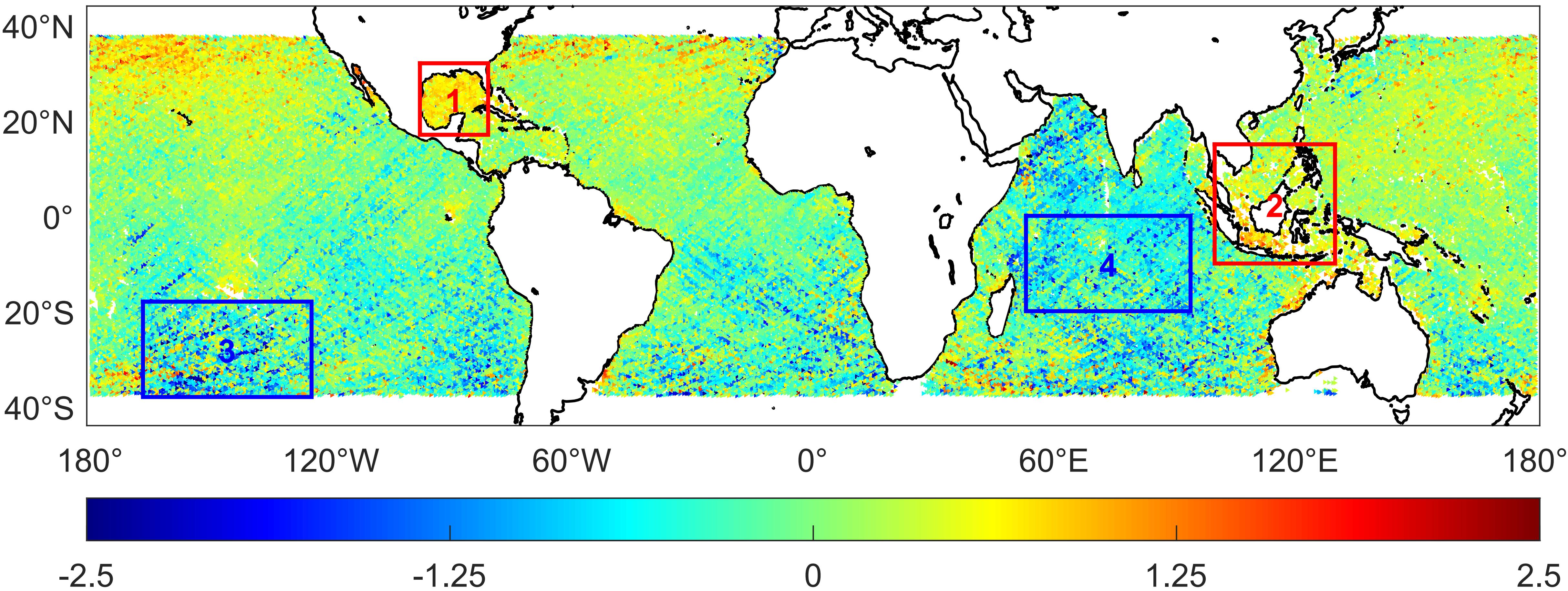}
        \centerline{(f)}
    \end{minipage}
    \hfill
    \begin{minipage}[t]{0.492\textwidth}
        \centering
        \includegraphics[width=\textwidth]{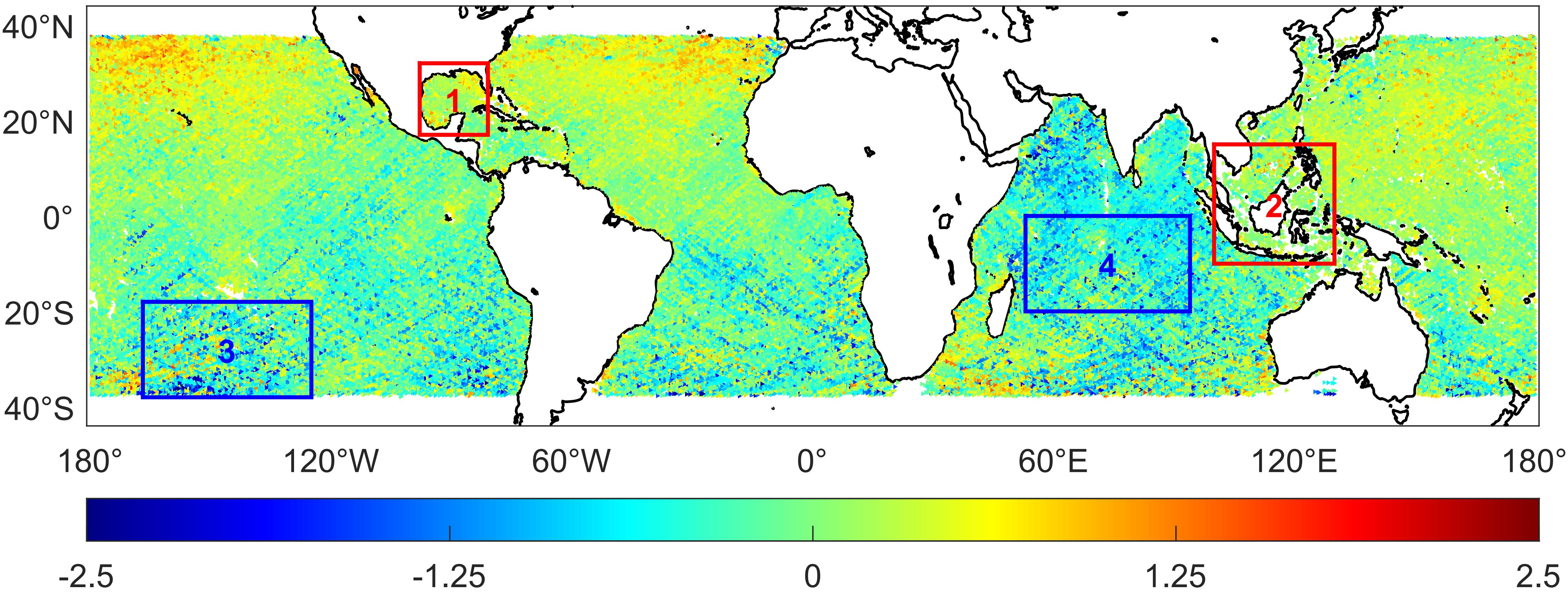}
        \centerline{(g)}
    \end{minipage}

    \vspace{0.5mm} 

    \begin{minipage}[t]{0.492\textwidth}
        \centering
        \includegraphics[width=\textwidth]{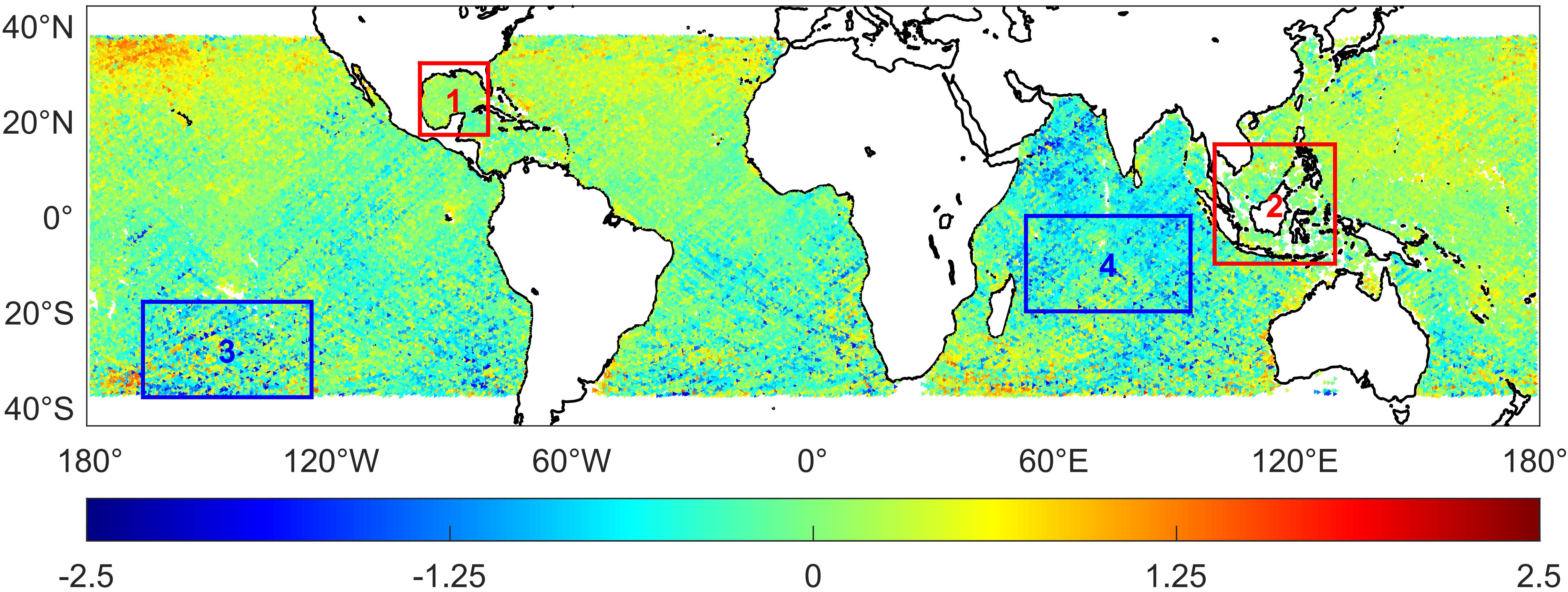}
        \centerline{(h)}
    \end{minipage}
    \hfill
    \begin{minipage}[t]{0.492\textwidth}
        \centering
        \includegraphics[width=\textwidth]{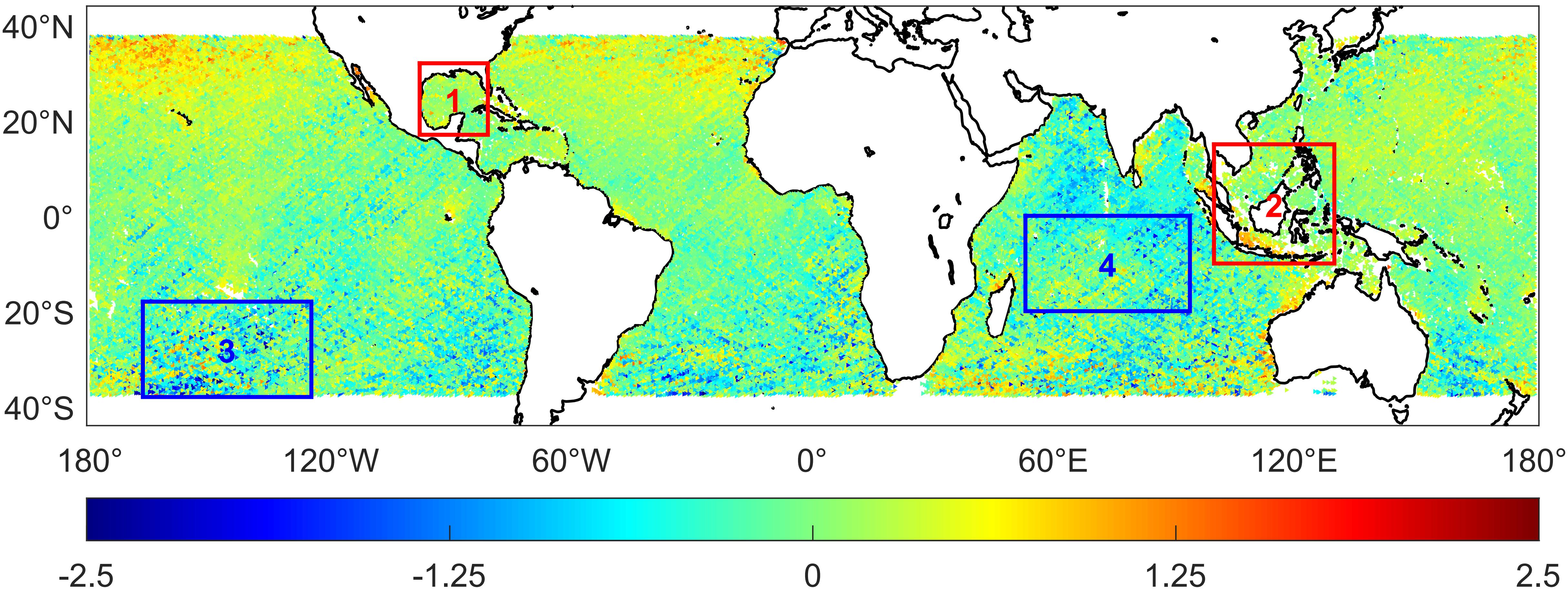}
        \centerline{(i)}
    \end{minipage}
    
    \caption{Global distribution of ERA5 SWH and Bias: (a) ERA5 SWH distribution, (b)-(i) Bias distribution of different models - (b) CNN-ConvLSTM, (c) ViT-Wave, (d) WaveTransNet, (e) $\text{MLP}_f$-CD, (f) $\text{CNN}_f$-CD, (g) SCAWaveNet-CI, (h) SCAWaveNet-CD, (i) SCAWaveNet-CD-wind.}
    
    \label{fig:global_swh_bias}
\end{figure*}

In addition to quantitative results, we conducted qualitative experiments to provide a more intuitive assessment of model performance. Regarding the prediction trends in the test set, Fig.~\ref{fig:testset_scatter} presents scatter density plots for the models on channel 1, where the x-axis represents the ERA5 SWH and the y-axis denotes the predicted SWH. The black dashed line corresponds to the ideal case indicating perfect agreement between predictions and references, while the pink line denotes the regression fit, highlighting the overall trend of the scatter distribution. The color bar on the right indicates the density scale, ranging from $10^{0}$ to $10^{4}$. 

As shown in Fig.~\ref{fig:testset_scatter}(g), the fitted line of SCAWaveNet-CD is closest to the ideal line compared to other models, especially in the 0–1 m range, consistent with the findings in Fig.~\ref{fig:diff_range}. While SCAWaveNet-CI (Fig.~\ref{fig:testset_scatter}(f)) also performs well at low SWH, it tends to underestimate SWH when SWH \textgreater 6 m. SCAWaveNet-CD exhibits fewer underestimations, suggesting that the incorporation of channel attention effectively alleviates the high-SWH underestimation. WaveTransNet (Fig.~\ref{fig:testset_scatter}(c)) and $\text{CNN}_f$-CD (Fig.~\ref{fig:testset_scatter}(e)) exhibit similar scatter distributions. This indicates that spatial attention in WaveTransNet contributes to effective feature extraction, and channel information interaction further enhances the representational capacity of CNNs. The fitted line of $\text{MLP}_f$-CD (Fig.~\ref{fig:testset_scatter}(d)) deviates from the ideal line, which exhibits the most severe underestimation. This implies that MLPs capture highly localized features in spatial and channel dimensions. ViT-Wave (Fig.~\ref{fig:testset_scatter}(b)), which extracts features only from DDMs without processing APs, exhibits poor performance, particularly at high SWH. In contrast, CNN-ConvLSTM (Fig.~\ref{fig:testset_scatter}(a)), which uses MLP for APs feature extraction, has a fitted line closer to the ideal line than that of ViT-Wave. This further demonstrates the importance of APs feature extraction. 

In terms of global retrieval errors, our algorithm also demonstrates superiority, as shown in Fig.~\ref{fig:global_swh_bias}. Fig.~\ref{fig:global_swh_bias}(a) illustrates the global distribution of the ERA5 test set. ERA5 SWH exhibits a hemispheric difference: lower values are generally observed in the Northern Hemisphere, whereas higher values prevail in the Southern Hemisphere, and increase with latitude. Additionally, the northwestern Indian Ocean in the Northern Hemisphere has relatively higher SWH values due to the influence of monsoonal currents. Figs.~\ref{fig:global_swh_bias}(b)-(i) present the global bias distribution between model predictions and ERA5 values. To further examine the spatial patterns of the bias, four key regions are selected for detailed analysis: two with potential overestimation (Regions 1 and 2) and two with underestimation (Regions 3 and 4). Region 1 covers the Gulf of Mexico and the northwestern Caribbean Sea, Region 2 includes the South China Sea, the Philippine Sea, and the Coral Sea, Region 3 is located in the central South Pacific Ocean and Region 4 lies in the central-eastern Indian Ocean. 

As shown in Fig.~\ref{fig:global_swh_bias}(h), SCAWaveNet-CD demonstrates superior performance across all regions compared to other models. In Region 3, both overestimation and underestimation are present, while Region 4 shows a slight underestimation. Predictions in Regions 1 and 2 align closely with the reference values. In contrast, Fig.~\ref{fig:global_swh_bias}(g) reveals that SCAWaveNet-CI slightly overestimates in Region 1 and significantly underestimates in Regions 3 and 4, which means the channel attention mechanism improves retrieval accuracy by reducing overestimation and underestimation areas. Figs.~\ref{fig:global_swh_bias}(d) and (f) display the results of WaveTransNet and $\text{CNN}_f$-CD, respectively. The two models yield similar results, with $\text{CNN}_f$-CD exhibiting more underestimation regions, while WaveTransNet displays more overestimated areas. Figs.~\ref{fig:global_swh_bias}(b), (c) and (e) depict the bias distributions of CNN-ConvLSTM, ViT-Wave, and $\text{MLP}_f$-CD, respectively. ViT-Wave, and $\text{MLP}_f$-CD show severe underestimation in Regions 3 and 4, consistent with the findings from Fig.~\ref{fig:testset_scatter}. Meanwhile, CNN-ConvLSTM shows significant overestimation in Regions 1 and 2, and this large prediction deviation also aligns with the results in Fig.~\ref{fig:diff_range}.

\begin{table}[t!]
\caption{Model Average Performance Comparision of SCAWaveNet-CD with different model configurations}
\label{diff_config}
\centering
\setlength{\tabcolsep}{5.5pt} 
\renewcommand{\arraystretch}{1.5}
\begin{tabular}{cccccccccc}
\toprule
\multirow{2}{*}{DDM} & \multicolumn{3}{c}{\multirow{1}{*}{AP}} & \multirow{2.2}{*}{\shortstack{RMSE\\(m)}} & \multirow{2.2}{*}{\shortstack{MAE\\(m)}} & \multirow{2.2}{*}{\shortstack{Bias\\(m)}} & \multirow{2.2}{*}{\shortstack{CC}} & \multirow{2.2}{*}{\shortstack{MAPE\\(\%)}} \\

\cline{2-4}
& \multirow{1.4}{*}{M} & \multirow{1.4}{*}{R} & \multirow{1.4}{*}{G} & & & & & & \\

\midrule
\checkmark & & & & 0.526 & 0.383 & -0.110 & 0.587 & 19.621 \\
& \checkmark &  &  & 0.492 & 0.368 & 0.017 & 0.636 & 18.569 \\
&  & \checkmark &  & 0.545 & 0.397 & 0.092 & 0.554 & 20.385 \\
&  &  & \checkmark & 0.508 & 0.376 & 0.057 & 0.627 & 18.821 \\
& \checkmark & \checkmark &  & 0.469 & 0.362 & 0.035 & 0.661 & 17.994 \\
& \checkmark &  & \checkmark & 0.458 & 0.352 & 0.030 & 0.686 & 17.762 \\
&  & \checkmark & \checkmark & 0.481 & 0.403 & 0.045 & 0.649 & 18.186 \\
& \checkmark & \checkmark & \checkmark & 0.452 & 0.349 & 0.027 & 0.698 & 17.529 \\
\checkmark & \checkmark & \checkmark & \checkmark & \textbf{0.438} & \textbf{0.330} & \textbf{-0.012} & \textbf{0.716} & \textbf{16.274} \\

\bottomrule
\end{tabular}
\end{table}

\subsubsection{Ablation Experiments}
To demonstrate the effectiveness of each component in SCAWaveNet, Table~\ref{diff_config} compares the performance of SCAWaveNet-CD across different configurations. In the table, ``DDM" refers to the DDM branch and``AP" refers to the AP branch. Within the ``AP" category, ``M”, ``R” and ``G” correspond to DDM-related, Receiver-related, and Geometry-related APs, respectively. The results indicate that the model integrating both branches outperforms single-branch configurations, confirming the complementary roles of these branches in enhancing model performance. Specifically, when using only the DDM branch, the model achieves an average RMSE of 0.526 m; in contrast, incorporating the AP branch reduces this RMSE to 0.452 m, indicating that the AP branch contributes more significantly to performance improvement. Further analysis of the AP branch reveals that DDM-related APs yield the most substantial contribution to performance enhancement, followed by Geometry-related APs and finally Receiver-related APs. Moreover, combining multiple types of AP leads to additional performance gains, demonstrating the benefit of incorporating all three types of AP into the modeling process.

\subsubsection{Model Parameters and Complexity Analysis}
In addition to quantitative and qualitative analyses of model performance, we further investigate model parameters and computational complexity. As shown in Table~\ref{tab:Model_paramete_comparison}, ``Params" denotes the total number of model parameters, including all weights and biases in the network, and “FLOPs” denotes the number of floating-point operations required for a single forward pass, which quantifies the computational complexity of the model. Among four-channel models, SCAWaveNet achieves the smallest parameter count and the lowest FLOPs. Although SCAWaveNet-CD has approximately double the parameter count of SCAWaveNet-CI due to channel-wise interactions, their FLOPs remain comparable, indicating that SCAWaveNet-CD maintains favorable computational efficiency. For single-channel models, WaveTransNet yields the lowest parameter count and FLOPs. However, SCAWaveNet-CI outperforms WaveTransNet in both parameter count and complexity, while SCAWaveNet-CD achieves comparable FLOPs but superior overall performance. These results underscore the advantage of SCAWaveNet in balancing model performance, parameter count, and computational complexity.

\begin{table}[t!]
\caption{parameter and complexity comparison of all the models}
\label{tab:Model_paramete_comparison}
\centering
\setlength{\tabcolsep}{14pt} 
\renewcommand{\arraystretch}{1.3}
\begin{tabular}{c c c}
\toprule
 Model & Params(M)~$\downarrow$ & FLOPs(G)~$\downarrow$ \\
\midrule

$\text{MLP}_s$ & 2.075 & 12.156 \\
$\text{CNN}_s$ & 3.441 & 19.173 \\
Conv-BiLSTM & 5.502 & 37.459 \\
CNN-ConvLSTM & 2.241 & 13.584 \\
ViT-Wave & 7.075 & 49.310 \\
WaveTransNet & 0.928 & 8.891 \\
$\text{MLP}_f$-CI & 2.328 & 13.772 \\
$\text{MLP}_f$-CD & 4.815 & 32.198 \\
$\text{CNN}_f$-CI & 3.545 & 20.585 \\
$\text{CNN}_f$-CD & 12.028 & 73.694 \\
$\text{SCAWaveNet}$-CI & 0.915 & 7.721 \\
$\text{SCAWaveNet}$-CD & 1.966 & 9.812 \\

\bottomrule
\end{tabular}
\end{table}

\subsection{NDBC Buoy Data Evaluation}
\begin{figure*}[!ht]
    \centering
    \includegraphics[width=\linewidth]{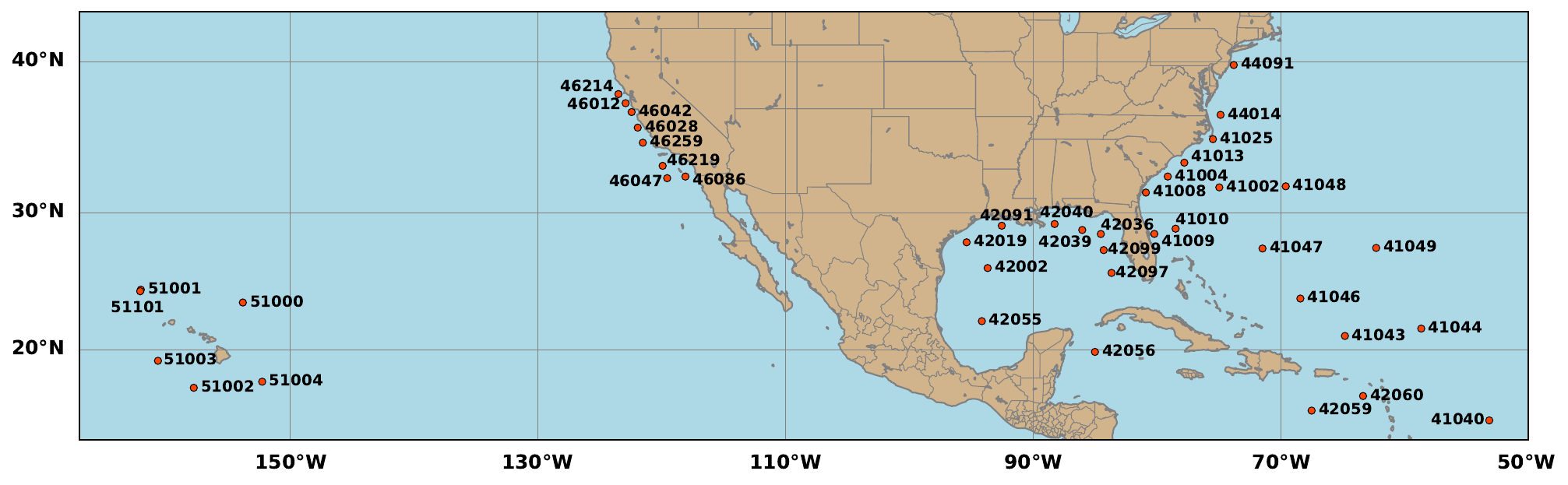}
    \caption{Spatial distribution of the selected NDBC buoy stations.}
    \label{fig:buoy stations}
\end{figure*}

The Buoy data serve as a critical source of ocean observations. Unlike model-generated data, buoy measurements are continuously collected via onboard sensors and instruments, offering higher accuracy and better representation of real-world marine conditions. Therefore, buoy-derived SWH measurements are widely regarded as a reliable reference for ocean state assessment, and it is necessary to evaluate model's generalization and robustness with buoy data. In this study, NDBC buoy data are utilized for model evaluation. Specifically, 42 NDBC buoys were selected, as illustrated in Fig.~\ref{fig:buoy stations}, where red dots mark their locations, predominantly distributed along the eastern and western coasts of America and near the Hawaiian Islands. These data align with the temporal range of the test set. Data from each channel are individually matched based on close temporal and spatial proximity, and then merged into four-channel samples, yielding a total of 4,438 four-channel data samples (17,752 data points).

\begin{figure*}[!t]
    \centering
    \begin{minipage}{0.49\columnwidth}
        \centering
        \includegraphics[width=\linewidth]{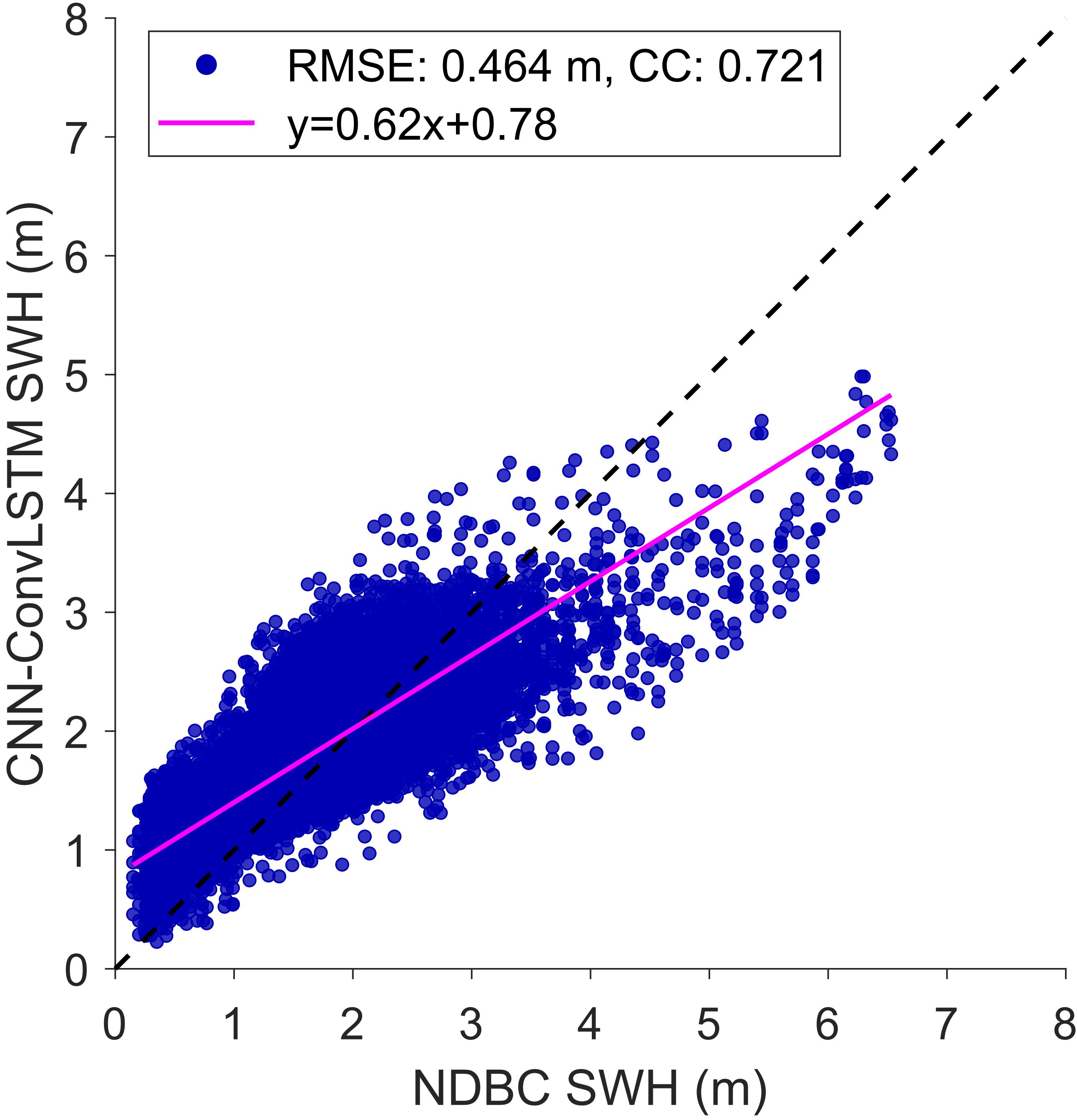}
        \centerline{(a)}
    \end{minipage}\label{subfig:buoy_ConvLSTM}
    \hfil
    \begin{minipage}{0.49\columnwidth}
        \centering
        \includegraphics[width=\linewidth]{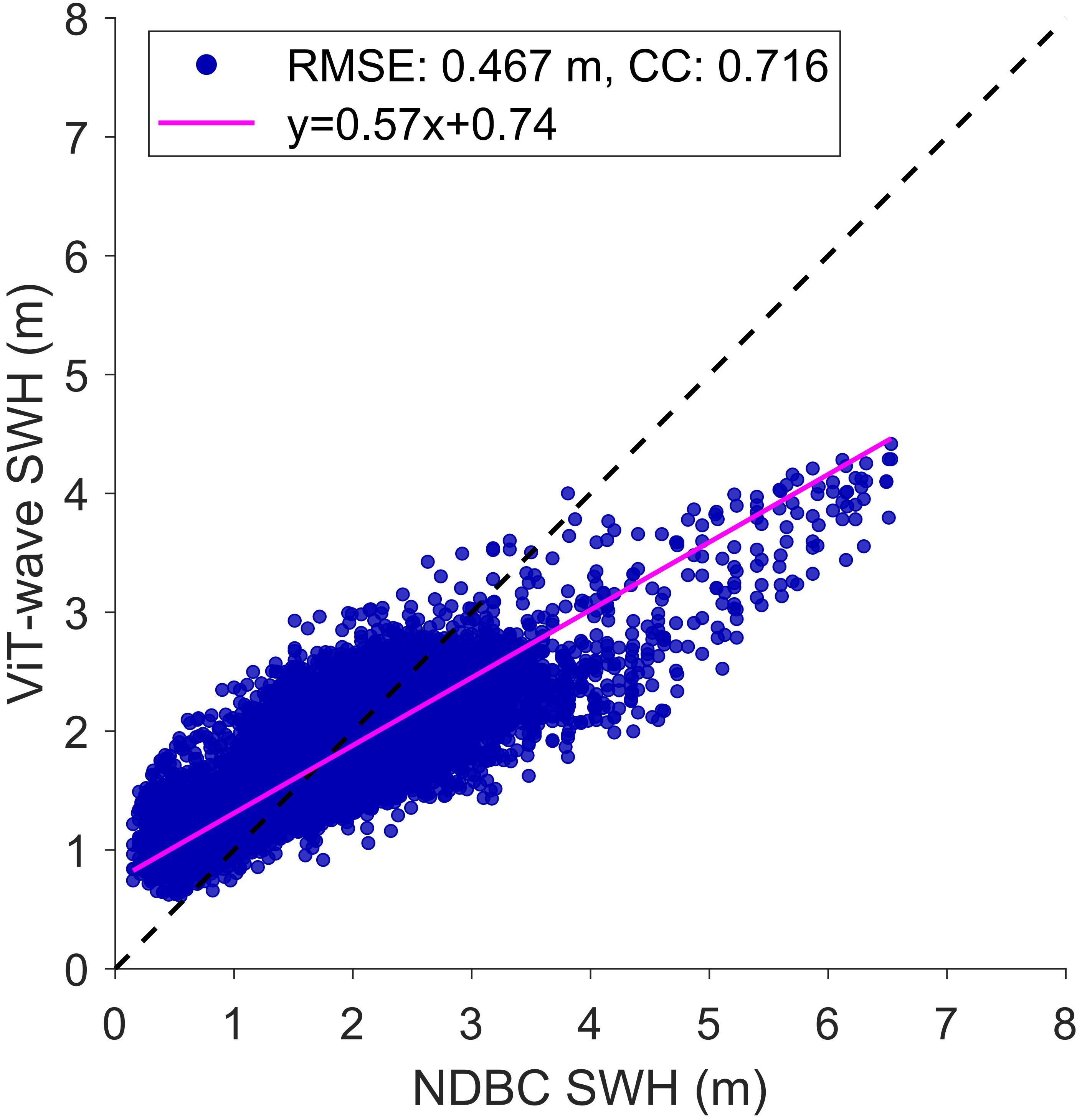}
        \centerline{(b)}
    \end{minipage}\label{subfig:buoy_ViT}
    \hfil
    \begin{minipage}{0.49\columnwidth}
        \centering
        \includegraphics[width=\linewidth]{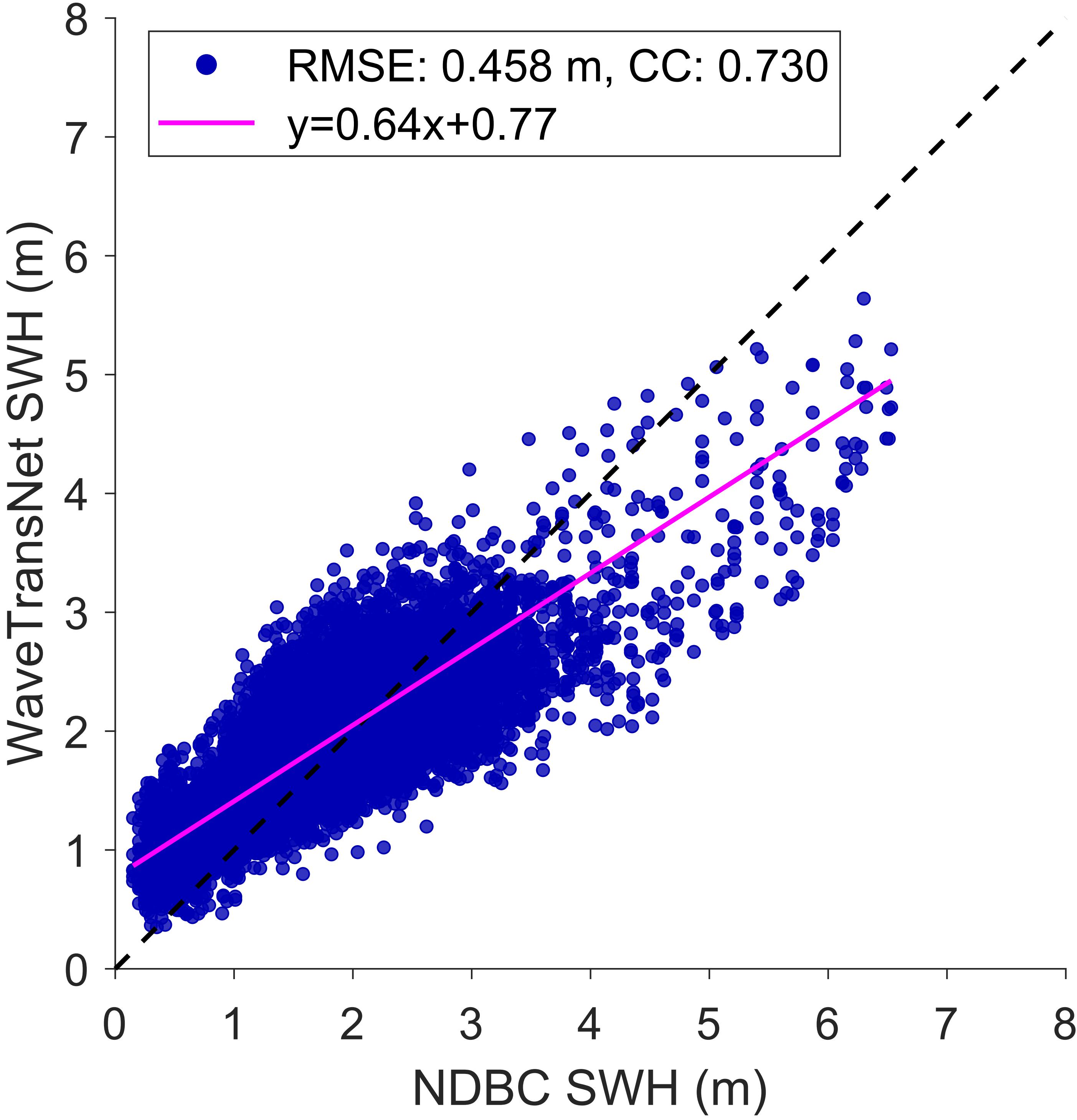}
        \centerline{(c)}
    \end{minipage}\label{subfig:buoy_WaveTransNet}
    \hfil
    \begin{minipage}{0.49\columnwidth}
        \centering
        \includegraphics[width=\linewidth]{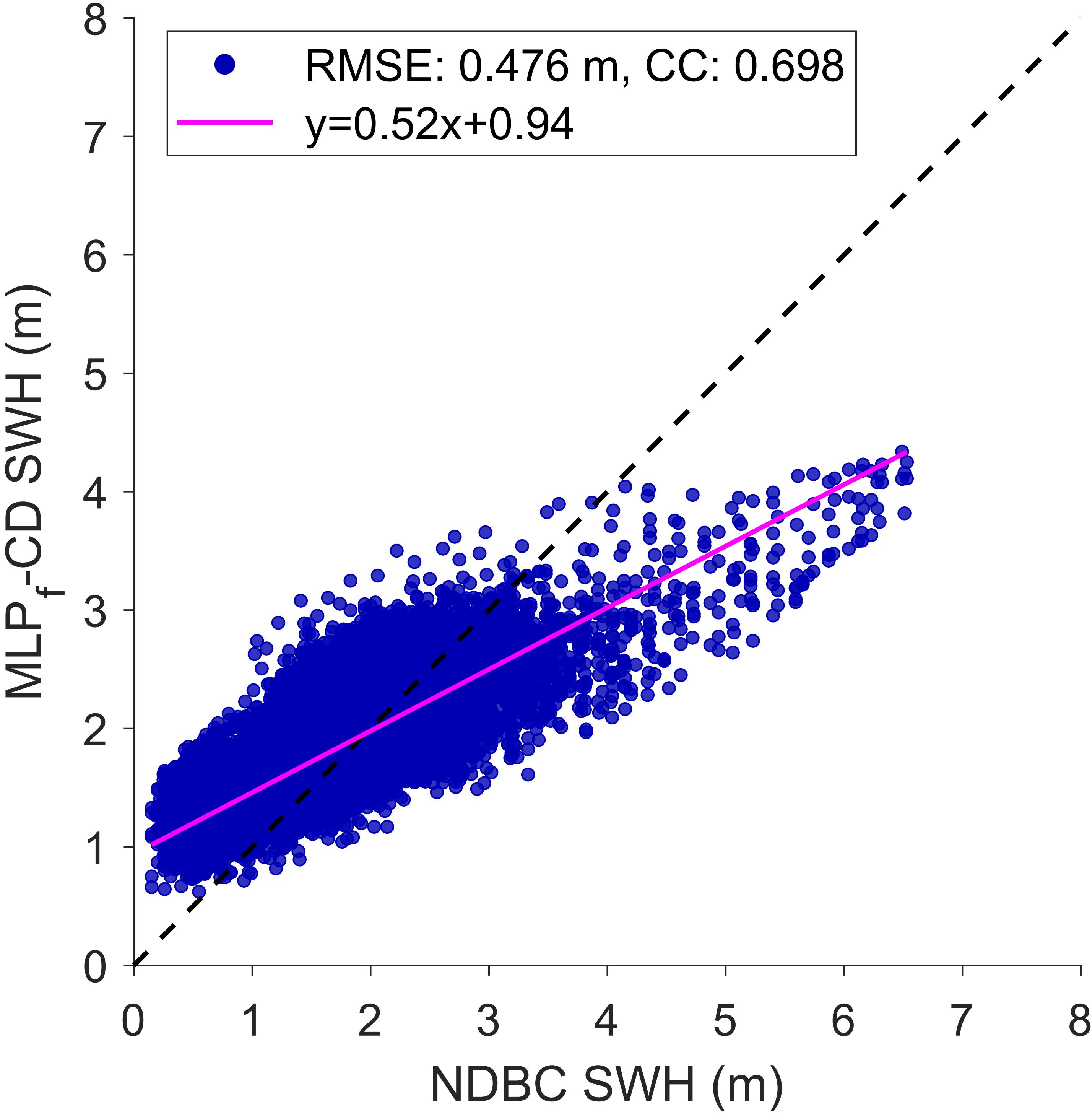}
        \centerline{(d)}
    \end{minipage}\label{subfig:buoy_MLP_f-CD}

    \begin{minipage}{0.49\columnwidth}
        \centering
        \includegraphics[width=\linewidth]{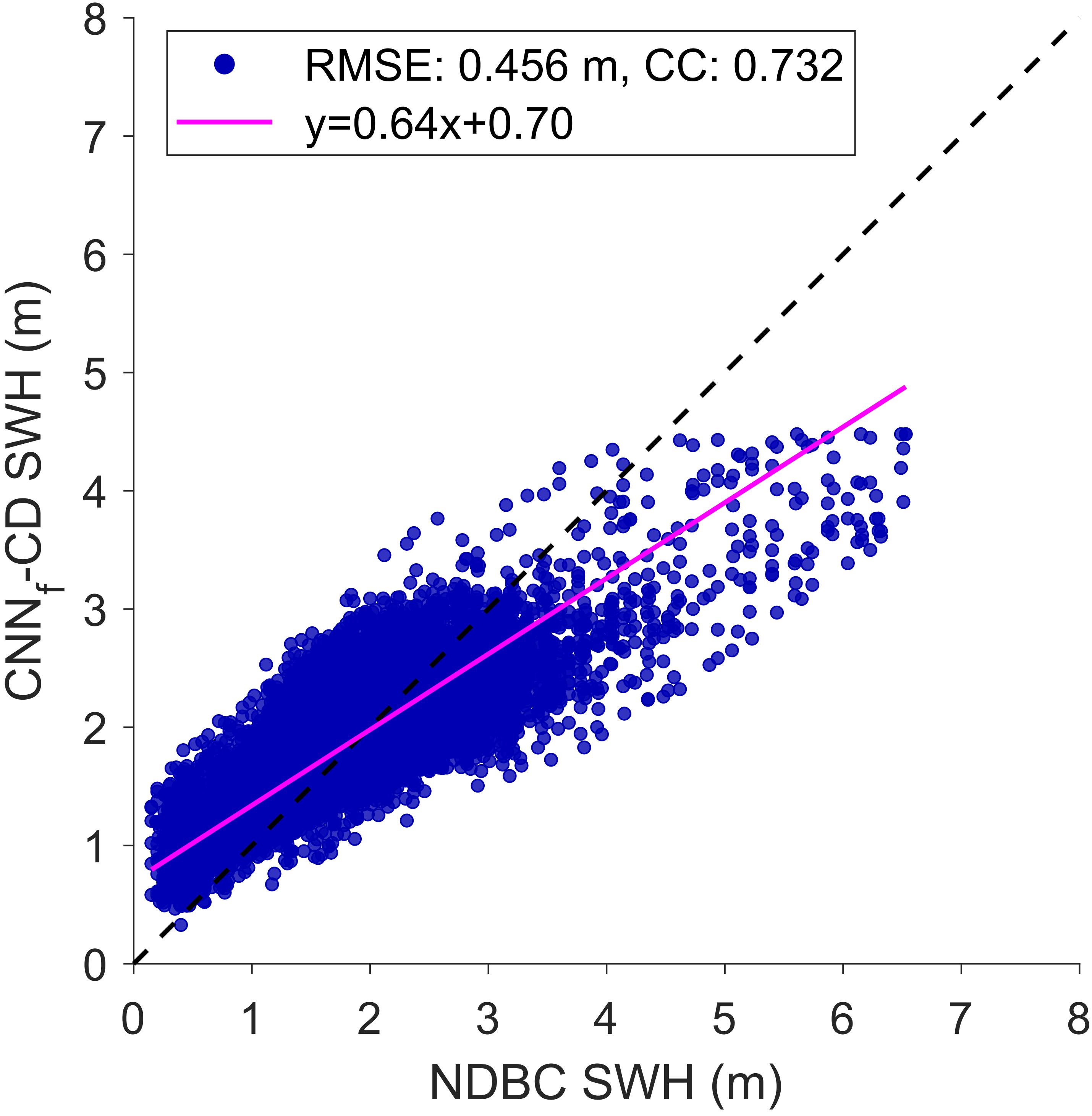}
        \centerline{(e)}
    \end{minipage}\label{subfig:buoy_CNN_f-CD}
    \hfil
    \begin{minipage}{0.49\columnwidth}
        \centering
        \includegraphics[width=\linewidth]{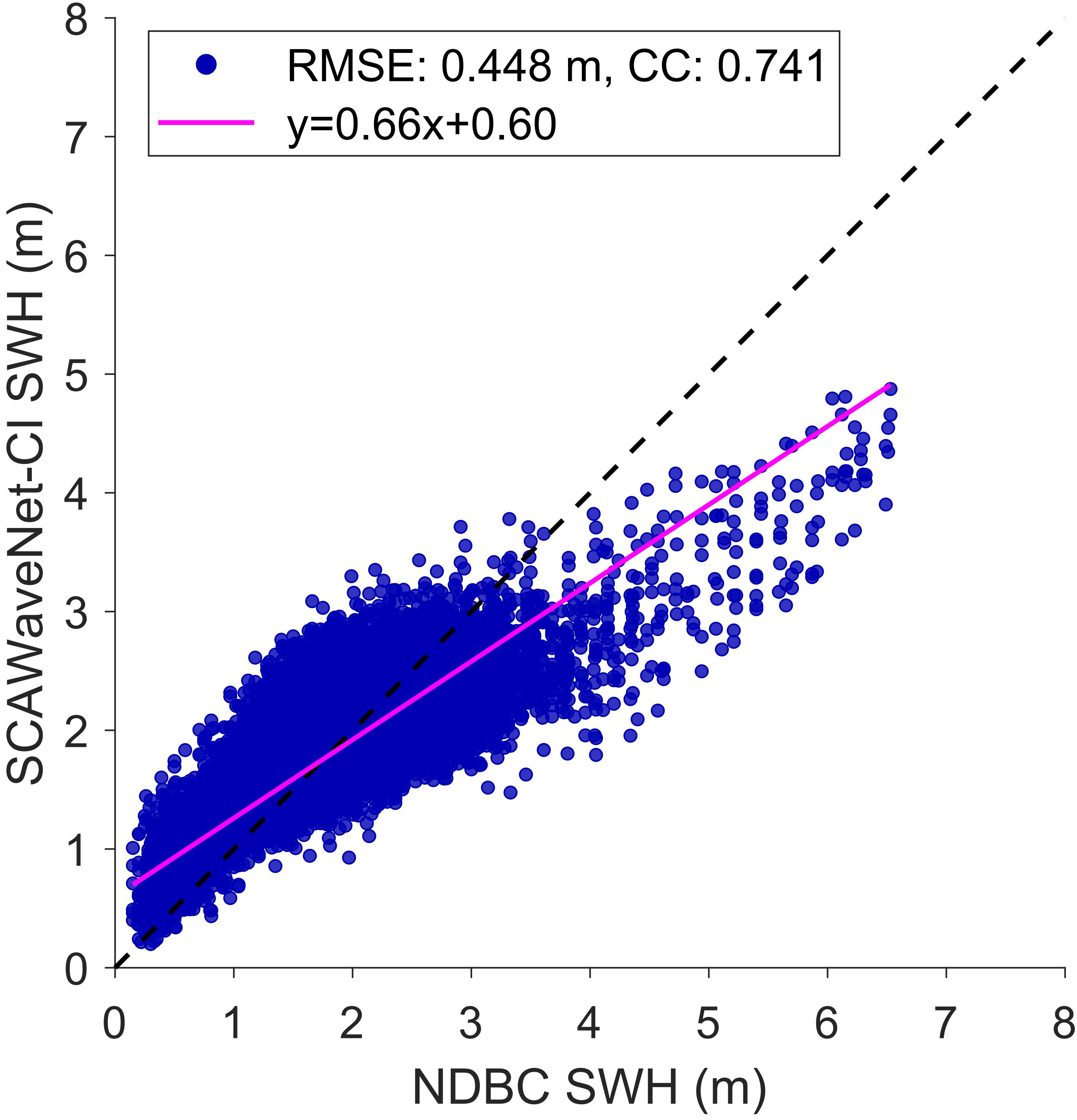}
        \centerline{(f)}
    \end{minipage}\label{subfig:buoy_SCAWaveNet-CI}
    \hfil
    \begin{minipage}{0.49\columnwidth}
        \centering
        \includegraphics[width=\linewidth]{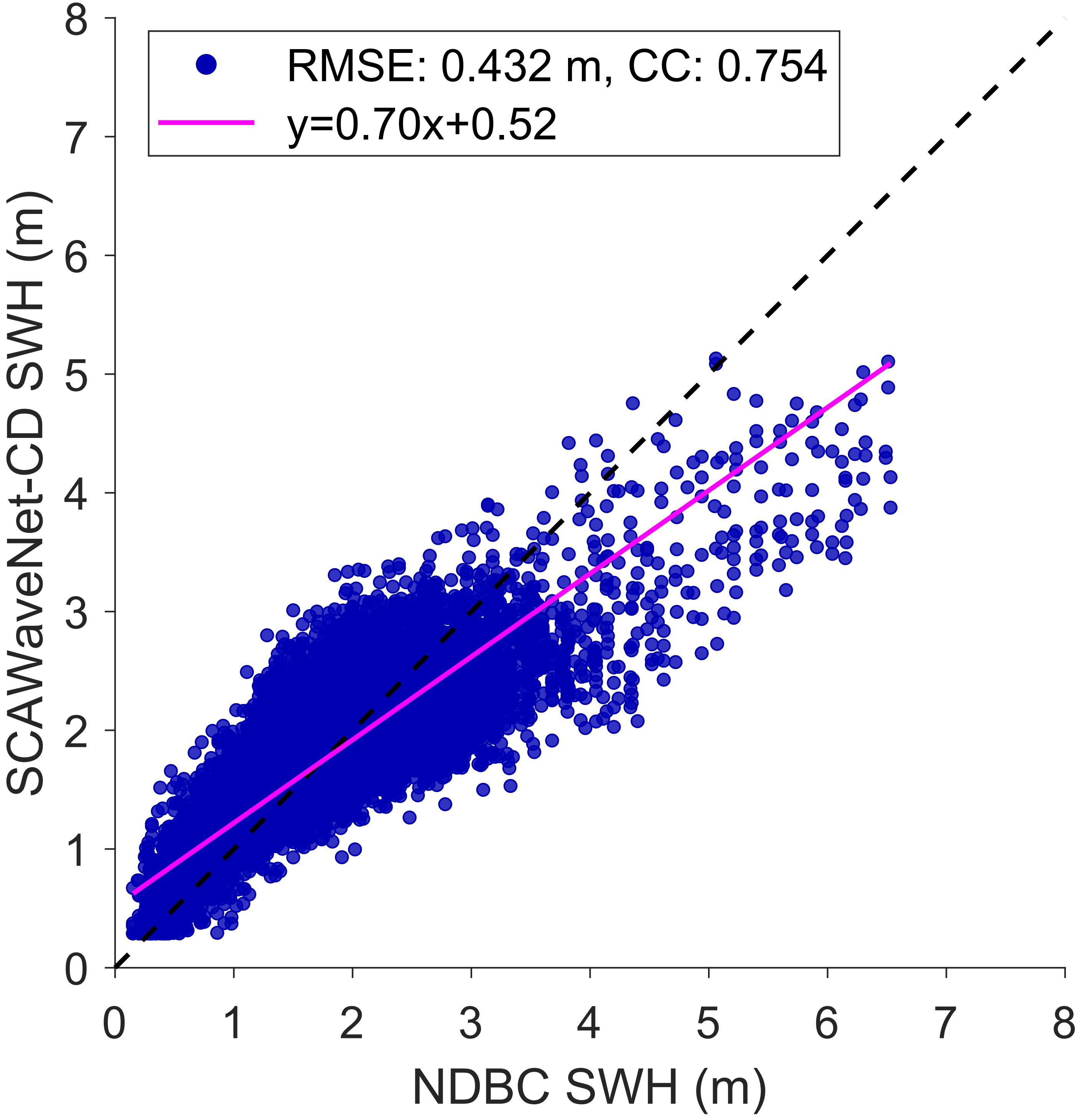}
        \centerline{(g)}
    \end{minipage}\label{subfig:buoy_SCAWaveNet-CD}
    \hfil
    \begin{minipage}{0.49\columnwidth}
        \centering
        \includegraphics[width=\linewidth]{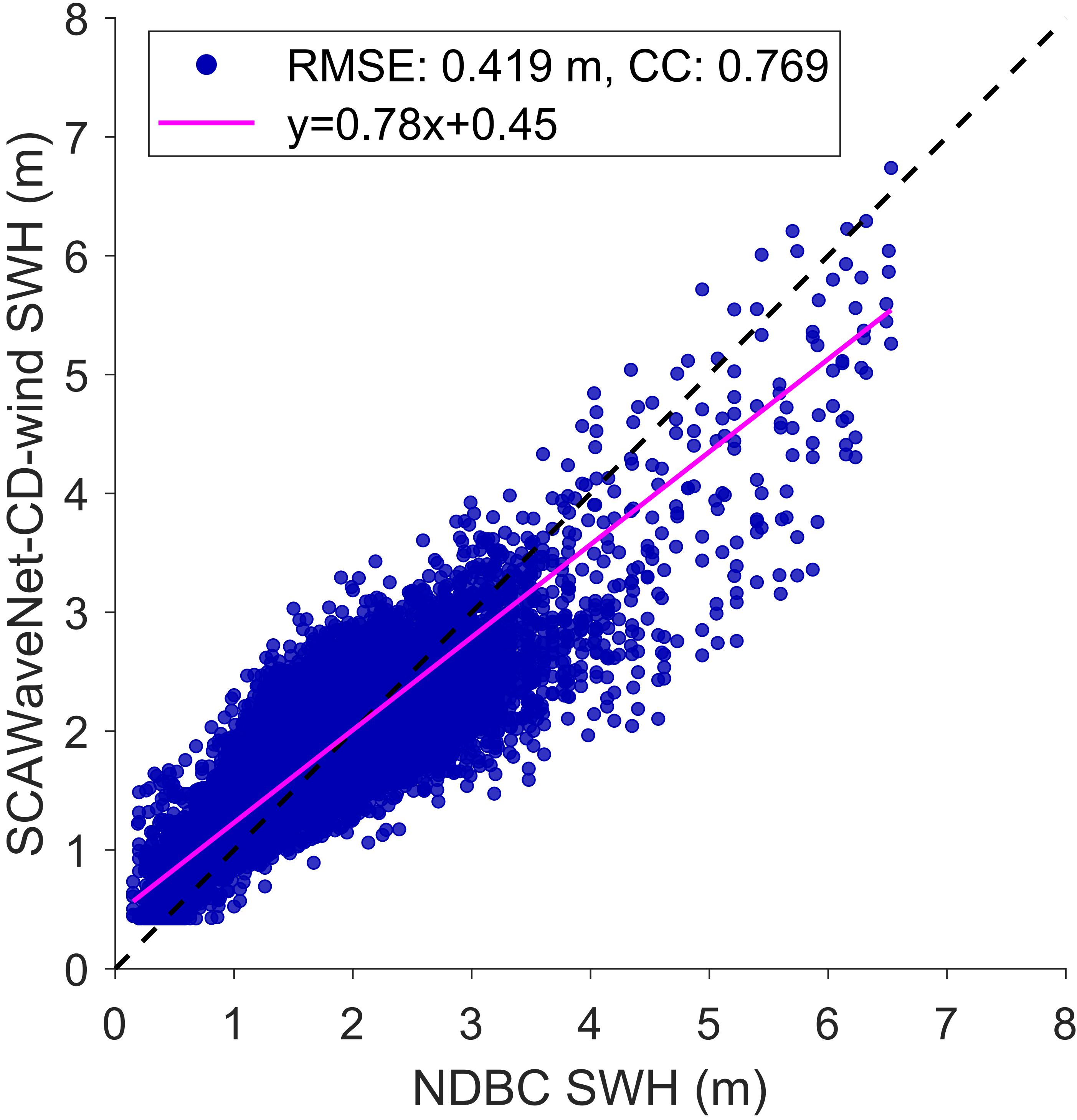}
        \centerline{(h)}
    \end{minipage}\label{subfig:buoy_SCAWaveNet-CD-wind}

    \caption{Scatter plots of model predictions for CYGNSS-Buoy dataset: (a) CNN-ConvLSTM, (b) ViT-Wave, (c) WaveTransNet, (d) $\text{MLP}_f$-CD, (e) $\text{CNN}_f$-CD, (f) SCAWaveNet-CI, (g) SCAWaveNet-CD, (h) SCAWaveNet-CD-wind.}
    \label{fig:buoy_scatter}
\end{figure*}

\begin{table}[t!]
\caption{Performance comparison of models on CYGNSS-Buoy data}
\label{tab:buoy_model_comparison}
\centering
\setlength{\tabcolsep}{3pt} 
\renewcommand{\arraystretch}{1.5}
\begin{tabular}{cccccccccc}
\toprule
\multirow{1}{*}{{Model}} & \multirow{1}{*}{\shortstack{RMSE(m)}} & \multirow{1}{*}{\shortstack{MAE(m)}} & \multirow{1}{*}{\shortstack{Bias(m)}} & \multirow{1}{*}{\shortstack{CC}} & \multirow{1}{*}{\shortstack{MAPE(\%)}} \\

\midrule
\rowcolor[gray]{0.8} ERA5 & 0.213 & 0.154 & -0.005 & 0.954 & 9.712 \\
$\text{MLP}_s$ & 0.490 & 0.383 & -0.128 & 0.683 & 20.034 \\
$\text{CNN}_s$ & 0.478 & 0.377 & -0.101 & 0.694 & 19.356 \\
Conv-BiLSTM & 0.473 & 0.375 & 0.079 & 0.706 & 18.827 \\
CNN-ConvLSTM & 0.464 & 0.366 & 0.069 & 0.721 & 18.171 \\
ViT-Wave & 0.467 & 0.368 & -0.077 & 0.716 & 18.465 \\
WaveTransNet & 0.458 & 0.357 & 0.045 & 0.730 & 17.693 \\
$\text{MLP}_f$-CI & 0.488 & 0.380 & -0.121 & 0.685 & 19.885 \\
$\text{MLP}_f$-CD & 0.476 & 0.373 & -0.104 & 0.698 & 19.025 \\
$\text{CNN}_f$-CI & 0.475 & 0.375 & -0.90 & 0.702 & 18.953 \\
$\text{CNN}_f$-CD & 0.456 & 0.353 & 0.039 & 0.732 & 17.621 \\
$\text{SCAWaveNet}$-CI & 0.448 & 0.345 & -0.036 & 0.741 & 16.749 \\
$\text{SCAWaveNet}$-CD & \underline{0.432} & \underline{0.337} & \underline{0.023} & \underline{0.754} & \underline{15.962} \\
$\text{SCAWaveNet}$-CD-wind & \textbf{0.419} & \textbf{0.321} & \textbf{0.006} & \textbf{0.769} & \textbf{14.266} \\

\bottomrule
\end{tabular}
\end{table}

Due to variations in input format across different models, we adopted distinct strategies for model evaluation. For single-channel models, data from all channels were merged for evaluation. For four-channel models, we used four-channel data to test the model and then flattened the prediction by channel to calculate the final results. The evaluation results of all models are shown in Table~\ref{tab:buoy_model_comparison}. As a global reanalysis dataset, ERA5 exhibits high consistency with the NDBC buoy data, with an average CC of 0.954 across four channels. This result confirms the reliability of using ERA5 as a reference for model evaluation. Among the evaluated results, SCAWaveNet-CD achieves the best performance on buoy data, with an average RMSE of 0.432 m and a CC of 0.754. SCAWaveNet-CI closely follows, with an RMSE of 0.448 m and a CC of 0.741. These findings indicate that the SCAWaveNet architecture effectively captures meaningful features from buoy data, with channel attention further enhancing feature representation. In contrast, other models exhibit performance degradation, suggesting a domain gap between ERA5 and buoy data. Notably, WaveTransNet still performs well among single-channel models, consistent with the ERA5 evaluation results, highlighting the effectiveness of spatial attention for feature extraction from DDM and APs.  

Fig.~\ref{fig:buoy_scatter} shows the scatter density plots on the CYGNSS-Buoy test set. The models are consistent with those used in Fig.~\ref{fig:testset_scatter}. As shown in Fig.~\ref{fig:buoy_scatter}(g), the fitted line of SCAWaveNet-CD is closest to the ideal line, followed by SCAWaveNet-CI (Fig.~\ref{fig:buoy_scatter}(f)). WaveTransNet (Fig.~\ref{fig:buoy_scatter}(c)) and $\text{CNN}_f$-CD (Fig.~\ref{fig:buoy_scatter}(e)) demonstrate similar trends: $\text{CNN}_f$-CD exhibits a more concentrated distribution with evident underestimation, whereas WaveTransNet shows slightly greater dispersion but less underestimation. $\text{MLP}_f$-CD (Fig.~\ref{fig:buoy_scatter}(d)) and ViT-Wave (Fig.~\ref{fig:buoy_scatter}(b)) display greater deviations from the ideal line, and suffer from severe underestimation. In summary, SCAWaveNet demonstrates robust performance on the buoy dataset. Although other models exhibit performance degradation, their prediction trends remain consistent with those observed in the ERA5 evaluations.

\subsection{Effect of wind speed on model performance}
In previous experiments, SCAWaveNet-CD demonstrated strong performance across various scenarios. However, a persistent underestimation trend for high SWH levels indicates potential for further improvement. Wind speed is highly correlated with SWH and plays a crucial role in SWH retrieval\cite{qiao2024wavetransnet, wang2022retrieval}. Therefore, it is necessary to examine the impact of wind speed on the proposed model. To this end, we integrated the ERA5 synthetic wind speed into SCAWaveNet-CD, creating an enhanced variant named SCAWaveNet-CD-wind.

Table~\ref{tab:w/o wind} presents the overall performance of the two models on ERA5 data. It can be seen that wind speed significantly improves model performance. Compared to SCAWaveNet-CD, SCAWaveNet-CD-wind achieves an average RMSE of 0.427 m and a CC of 0.743, representing relative improvements of 2.51\% (0.438 m) and 3.77\% (0.716). As detailed in Table~\ref{tab:buoy_model_comparison}, SCAWaveNet-CD-wind reaches an average RMSE of 0.419 m and a CC of 0.769, reflecting relative gains of 3.01\% (0.432 m) and 2\% (0.754), respectively. These improvements are further visualized in Fig.~\ref{fig:testset_scatter}(g)-(h) and Fig.~\ref{fig:buoy_scatter}(g)-(h). After incorporating wind speed, the model shows a significant reduction in the underestimation for SWH \textgreater 4 m, with predictions aligning more closely with the ideal line. In addition, a comparison of Figs.~\ref{fig:global_swh_bias}(h) and (i) reveals that the underestimation in Regions 3 and 4 is substantially alleviated, with predictions becoming more consistent with the ERA5 values. These results confirm that integrating wind speed effectively enhances model performance by mitigating underestimation in high-SWH scenarios.

\begin{table}[t!]
\caption{Performance Comparison of SCAWaveNet-CD With and Without Wind Speed Integration}
\centering
\setlength{\tabcolsep}{3.6pt} 
\renewcommand{\arraystretch}{1.1}

\begin{tabular}{ccccccc}
\toprule

Model & Ch. & RMSE(m) & MAE(m) & Bias(m) & CC & MAPE(\%)\\

\midrule
 \multirow{5}{*}{\shortstack{SCAWaveNet-\\CD}} 
 & 1 & 0.440 & 0.333 & 0.019 & 0.713 & 16.509 \\
 & 2 & 0.438 & 0.330 & -0.017 & 0.715 & 16.396 \\
 & 3 & 0.437 & 0.329 & -0.016 & 0.717 & 16.158 \\
 & 4 & 0.436 & 0.328 & -0.010 & 0.719 & 15.932 \\
 & Avg. & 0.438 & 0.330 & -0.012 & 0.716 & 16.274\\
\midrule
\multirow{5}{*}{\shortstack{SCAWaveNet-\\CD-wind}} 
 & 1 & 0.429 & 0.324 & 0.016 & 0.743 & 15.658 \\
 & 2 & 0.427 & 0.323 & 0.011 & 0.743 & 15.549 \\
 & 3 & 0.426 & 0.320 & -0.016 & 0.743 & 15.291 \\
 & 4 & 0.425 & 0.319 & -0.018  & 0.744 & 15.223 \\
 & Avg. & \textbf{0.427} & \textbf{0.322} & \textbf{-0.002} & \textbf{0.743} & \textbf{15.455} \\
\bottomrule

\label{tab:w/o wind}

\end{tabular}
\end{table}

\section{Discussion} \label{Discussion}
This paper investigates channel information from CYGNSS data and provides a detailed analysis of SWH retrieval models. However, several important aspects remain to be discussed. 

\subsubsection{Impact of Input Format} Existing DL-based models lack discussion of input formats during the training process, especially when considering channel information. We investigate two training strategies. First, channel-specific training involves training four separate models, each using data from a single channel independently. Second, channel-general training merges data from all channels to train a single general model. To analyze the impact of these strategies on model performance, we applied both strategies to four representative single-channel models. The results (presented in the Appendix) show that channel-specific models exhibit better performance than channel-general models. Furthermore, channel-general models display performance variations across different channels, suggesting that they fail to learn balanced presentations for each channel. Therefore, we adopt the channel-specific strategy for experimental analysis.

\subsubsection{The choice of Channel Strategy} This paper introduces three channel strategies. However, only the effects of CD and CI were discussed in the experiments, while the CP strategy was not included. This is because CP involves more diverse and complex modeling approaches. Specifically, CP can be implemented either with a fixed number of related channels or a dynamic adaptation mechanism over time, which requires further investigation in future research.

\subsubsection{Impact of Other Environmental Factors} Beyond wind speed, other environmental factors such as rainfall, wind direction, sea surface temperature, and salinity also affect the model performance. Since this paper focuses on analyzing the impact of channel information on SWH modeling, only the influence of wind speed on SCAWaveNet is considered. Future research could further examine the effects of these factors on different modeling strategies.

\subsubsection{Data Imbalance} CYGNSS data exhibit a highly imbalanced distribution, with over 80\% of samples within the 1-3 m. Consequently, the model primarily learns features from these frequent cases. For rare samples (e.g., extreme SWH values), it tends to predict values close to the dominant SWH, leading to underestimation under extreme conditions. Although SCAWaveNet performs better than other models on rare samples, this issue still exists. Future work can explore specialized techniques to address this imbalance.

\section{Conclusion} 
\label{Conclusion}
Motivated by the data acquisition principles of CYGNSS, this study first applies channel information to the SWH retrieval task. Leveraging attention mechanisms for dynamic feature adaptation and long-range dependency modeling, we propose a spatial-channel attention-based network, namely SCAWaveNet, aimed at better extracting informative features from the data. Specifically, patch embeddings from each channel of DDMs are treated as independent attention heads to enable joint modeling of spatial and channel dependencies. Additionally, a lightweight attention module is applied to APs, which assigns attention weights along both spatial and channel dimensions. The final merged features integrate spatial and channel-level information from both DDMs and APs. Unlike single-channel models, SCAWaveNet employs a four-channel architecture, using data from all four channels as inputs and predicting SWH for each channel. This design incorporates richer observations and enables the simultaneous retrieval of four SWH values. Experimental results show that SCAWaveNet outperforms state-of-the-art models on four-channel ERA5 and buoy data, achieving superior accuracy and robustness. In particular, the model exhibits significant advantages in the 0–1 m range, with regional validation conducted on the Gulf of Mexico, South China Sea, Philippine Sea, and Coral Sea supporting this finding. We also explore the impact of different channel strategies on the four-channel model. The model using the CD strategy outperforms the CI strategy, highlighting the importance of channel information interaction. Furthermore, we examine the impact of wind speed on model performance. Wind speed improves model accuracy by reducing underestimation at high SWH values.

\section*{Acknowledgments}
The authors would like to thank NASA for providing the CYGNSS data, the European Centre for Medium-Range Weather Forecasts for the SWH product, and the National Data Buoy Center for the buoy observations. They are also grateful to all individuals and organizations whose insight, support, or assistance has contributed to the development of this work.

\nocite{*}
\bibliographystyle{IEEEtran}


\vfill

\end{document}